\documentclass[accepted,specialissue]{melba}

\usepackage[english]{babel}  
\usepackage{amsmath,amsfonts}

\usepackage[caption=false,font=scriptsize,labelfont=sf,textfont=sf]{subfig}
\usepackage{graphicx}
\usepackage{bm}

\usepackage[narkup=underlined]{changes}		

\newcommand{\ie}{\textit{i.e.}}
\newcommand{\eg}{\textit{e.g.}}

\melbaid{2025:050}  
\doi{10.59275/j.melba.2025-ab9a}
\melbaauthors{{\"O}zbulak, Jimenez-del-Toro, Fatoretto, Berton, and Anjos}  
\email{gokhan.ozbulak@idiap.ch}
\volume{3}
\firstpageno{938}  
\melbayear{2025}  
\datesubmitted{2025-04-21}  
\datepublished{2025-12-30}  

\melbaspecialissue{Special issue on Fairness of AI in Medical Imaging (FAIMI)}
\melbaspecialissueeditors{Veronika Cheplygina, Aasa Feragen, Andrew King, Ben Glocker, Enzo Ferrante, Eike Petersen, Esther Puyol-Ant{\'o}n, Melanie Ganz-Benjaminsen}

\ShortHeadings{A Multi-Objective Evaluation Framework for Analyzing Utility-Fairness Trade-Offs in Machine Learning Systems}{{\"O}zbulak, Jimenez-del-Toro, Fatoretto, Berton, and Anjos}

\title{A Multi-Objective Evaluation Framework for Analyzing Utility-Fairness Trade-Offs in Machine Learning Systems}

\author{
	\firstname G{\"o}khan \surname {\"O}zbulak\aff{1,2}\orcid{0000-0001-5568-1622},
	\firstname Oscar \surname Jimenez-del-Toro\aff{1}\orcid{0000-0002-0628-3764},
	\firstname Ma{\'i}ra \surname Fatoretto\aff{3}\orcid{0000-0002-2508-1586},
	\firstname Lilian \surname Berton\aff{3}\orcid{0000-0003-1397-6005},
	\firstname Andr{\'e} \surname Anjos\aff{1}\orcid{0000-0001-7248-4014}
}
\affiliations{
	\num 1 \addr Idiap Research Institute, Martigny, Switzerland \\
	\num 2 \addr \'{E}cole Polytechnique F\'{e}d\'{e}rale de Lausanne (EPFL), Lausanne, Switzerland \\
	\num 3 \addr Federal University of S{\~a}o Paulo (UNIFESP), S{\~a}o Paulo, Brazil
}

\abstract{
The evaluation of fairness models in Machine Learning involves complex challenges, such as defining appropriate metrics, balancing trade-offs between utility and fairness, and there are still gaps in this stage.
This work presents a novel multi-objective evaluation framework that enables the analysis of utility-fairness trade-offs in Machine Learning systems.
The framework was developed using criteria from Multi-Objective Optimization that collect comprehensive information regarding this complex evaluation task.
The assessment of multiple Machine Learning systems is summarized, both quantitatively and qualitatively, in a straightforward manner through a radar chart and a measurement table encompassing various aspects such as convergence, system capacity, and diversity.
The framework's compact representation of performance facilitates the comparative analysis of different Machine Learning strategies for decision-makers, in real-world applications, with single or multiple fairness requirements.
In particular, this study focuses on the medical imaging domain, where fairness considerations are crucial due to the potential impact of biased diagnostic systems on patient outcomes.
The proposed framework enables a systematic evaluation of multiple fairness constraints, helping to identify and mitigate disparities among demographic groups while maintaining diagnostic performance.
The framework is model-agnostic and flexible to be adapted to any kind of Machine Learning systems, that is, black- or white-box, any kind and quantity of evaluation metrics, including multidimensional fairness criteria.
The functionality and effectiveness of the proposed framework are shown with different simulations, and an empirical study conducted on three real-world medical imaging datasets with various Machine Learning systems.
Our evaluation framework is publicly available at \url{https://pypi.org/project/fairical}.
}

\keywords{Machine Learning, Multidimensional Fairness Evaluation, Multi-Objective Optimization, Utility-Fairness Trade-off, Medical Image Analysis}

\begin{document}

\twocolumn[\maketitle]

\section{Introduction}
\label{sec:intro}
The increasing integration of Machine Learning (ML) systems in day-to-day activities
offers significant opportunities, but it also raises critical concerns regarding
demographic fairness and
equity~\citep{xinying2023guide,pessach2022review,starke2022fairness,rabonato2025systematic}.
Fairness in ML pertains to the ethical imperative of ensuring that algorithms and
models do not discriminate or display bias, against individuals or groups based on
lawfully demographic attributes such as race, gender, age, and/or other
characteristics~\citep{barocas2023fairness}.
Fairness is a multi-faceted and complex concept with nuances directly linked to the
situation considered~\citep{castelnovo2022clarification,pessach2022review}.
Consequently, balancing and measuring multiple fairness criteria simultaneously, both
at a group and individual level, is a challenging task, resulting in different
definitions of fairness appropriate for different contexts~\citep{dutt2023fairtune}.

While Multi-Objective Optimization (MOO) provides a mathematical foundation for
balancing objectives in conflict, fairness in ML extends beyond conventional
optimization paradigms. Equity encompasses multiple, often conflicting notions, such as
group or individual fairness and equality of opportunity, that reflect distinct ethical
and social considerations. These criteria cannot be meaningfully represented by a
single objective function, as optimizing for one dimension of fairness may amplify
disparities in another. In this respect, ML based approaches that model and evaluate
the utility-fairness trade-off across multiple objectives can provide valuable insight
into how fairness manifests under different operational and demographic conditions.
However, as highlighted by~\cite{selbst2019fairness} in five traps, over-abstracting
fairness into purely mathematical formalizations can have a risk of detaching it from
the reality. Therefore, such analyses should be interpreted as structured,
assumption-bound explorations of fairness dynamics, rather than universal solutions to
the fairness problem in ML.

An unintended outcome when optimizing for a balanced treatment between genders is the
variation in predictive performance across other groups of demographic attributes. Many
techniques in real-world scenarios improve fairness at the expense of model utility,
with the minimum possible error of any fair classifier bounded by the difference in
base rates~\citep{zhao2022inherent}. This fundamental tension in algorithmic fairness
has been previously explored to improve the understanding of model bias and the limits
of artificial intelligence (AI)~\citep{wei2022fairness,wang2020towards}.
Model biases can be introduced through the optimization of certain objectives,
hyperparameter tuning, or simply due to the inherent characteristics of the datasets
used to train the models~\citep{yang2024limits}, such as biased training data, sampling
bias, label bias, exclusion, or historical bias. For example, if features correlated
with sensitive attributes (\eg, gender, race) are considered during training, the
models might latch onto these attributes, potentially resulting in demographically
biased outcomes.

The selection and modeling of fairness criteria are relatively new research directions
that require clear definitions on the mathematical expression of demographic equity.
Most ML approaches are currently evaluated without considering any fairness
criteria~\citep{akter2022glaucoma,lu2020multiobjective}. In recent years, some works
started using unique levels of fairness and utility, which fail to characterize ML
systems at every level of the utility-fairness trade-off, thus limiting subgroup and
intersectional evaluation~\citep{buolamwini2018gender}. The situation worsens when
multiple fairness and utility criteria come into play, as the fairness-aware
stakeholders and decision-makers may want to ensure, beyond utility, that multiple
fairness criteria are satisfied, such as race, gender, and age aiming to provide fair
ML services in production under different demographic settings.

In medical imaging, fairness challenges become more prominent because diagnostic
processes often involve multiple and conflicting clinical and operational objectives,
as ML models are increasingly used to support decisions in almost every field, such as
radiology, ophthalmology, and dermatology. Furthermore, variations in data acquisition
conditions, patient demographics (\eg, the deployment of diagnostic tools in a hospital
that tends to a heterogeneous community), and disease prevalence can lead to systematic
biases in model predictions. For instance, screening models for ophthalmology risk must
retain high sensitivity to avoid missed cases, while also ensuring that performance
does not systematically degrade for specific demographic subgroups. A representative
example arises in glaucoma, an eye disease that exhibits a higher prevalence among
Black populations, and within this group, male individuals show greater vulnerability
compared to females~\citep{khachatryan2019primary,luo2024harvard}. The high prevalence
of glaucoma contrasts with the scarcity of available data from the Black community,
contributing to disparities in model performance across racial groups and even between
genders within the same group. This mismatch highlights the necessity of evaluating ML
models not only for diagnostic utility but also for their equitable behavior across
different demographic attributes. Consequently, utility (diagnostic performance) and
multiple fairness constraints must be considered simultaneously. A principled way to
analyze such conflicting requirements is to employ multi-objective formalizations,
where each fairness criterion and the utility metric are treated as separate but
jointly optimized objectives rather than in isolation.

On the other hand, the development of ML systems addressing multiple objectives has
been extensively studied in the context of MOO systems. In such cases, performance is
evaluated by considering all possible trade-offs between the individual objectives,
resulting in an N-dimensional graph. This evaluation strategy can provide a basis to
better articulate user preferences in model comparison~\citep{gong2023influence}.
Although multi-objective measurements have been used in recent works to incorporate
single fairness constraints into developed models~\citep{little2023fairness}, to the
best of our knowledge, there are no frameworks that enable a comprehensive comparison
of ML systems under multiple utility and fairness criteria. In particular, multiple
fairness considerations based on MOO become even more critical in the context of
medical imaging. A framework that accounts for multiple fairness constraints
simultaneously is therefore essential to ensure equitable diagnostic performance across
different patient subgroups characterized by demographic attributes.

Conceptually, fairness in ML can be regarded as a multidimensional evaluation problem
rather than a single optimization objective. Instead of focusing on maximizing a
particular fairness metric, it is often more informative to examine how ML systems
perform across a spectrum of utility-fairness trade-offs. Such an analysis enables a
clearer understanding of both ideal cases, where equity across demographic groups is
maximized and practical situations in which certain fairness dimensions may be
compromised due to data, operational, or design constraints. Considering fairness in
this way allows researchers and practitioners to interpret model behavior within an
explicit space of trade-offs, rather than through isolated or averaged/maximized
performance scores. This perspective establishes the conceptual foundation for
evaluating fairness-aware ML systems in a structured, multi-objective manner, before
introducing any specific methodological framework.

In this work, we present an evaluation framework, supported by MOO principles, for the
challenging task of comparing ML systems under multiple utility-fairness trade-offs.
This approach allows the comparison of multiple systems in a common multidimensional
space, where multiple concurrent fairness criteria can come into play.
The framework's applicability is demonstrated through use cases based on medical
imaging, where fairness constraints play a crucial role in ensuring equitable
diagnostic performance.
Our contributions can be summarized as follows:
\begin{itemize}
    \item A model- and task-agnostic evaluation framework with a compact yet comprehensive
          representation, both qualitatively and quantitatively, of multiple utility-fairness
          trade-offs resulting from the deployment of ML systems, facilitating system performance
          analysis and comparison.
    \item The framework integrates multiple fairness metrics into the evaluation process,
          providing a more nuanced and multi-faceted assessment of model performance.
    \item A detailed analysis of the proposed framework and rationale through simulations of
          typical ML systems on synthetically generated data.
    \item An empirical study based on three real-world medical imaging datasets demonstrating the
          effectiveness of the proposed framework.
    \item An open-source implementation of the framework allowing reproduction of results
          established in this article, and further
          reuse\footnote{\url{https://pypi.org/project/fairical}}.
\end{itemize}

While the framework is demonstrated in the context of medical imaging, its formulation
is model-, metric-, and domain-agnostic, and can be readily applied to other
high-stakes ML systems (\eg, decision-support in finance, hiring, criminal justice, or
homeland security applications for biometrics) where multiple fairness and utility
objectives may be in conflict. By providing a unified view of utility-fairness
trade-off, the framework establishes a generalizable flow for cross-domain benchmarking
and transparent model selection across diverse application areas.

The paper is organized as follows: Section~\ref{sec:background} reviews previous works
tackling the evaluation of utility-fairness trade-off systems. Section~\ref{sec:method}
thoroughly describes the proposed evaluation framework, with a set of use cases and MOO
principles behind it. The applicability of the framework in real-world scenarios is
shown in Section~\ref{sec:empirical}, with the analysis of ML systems for three medical
imaging tasks. Finally, a discussion and conclusion with the key points and limitations
from this study are presented in Section~\ref{sec:discussion} and
Section~\ref{sec:conclusion}.

\section{Background and Related Work}
\label{sec:background}
Fairness in machine learning can be categorized based on criteria, sources of bias,
perspectives, methodologies, and trade-offs. Fairness criteria include demographic
parity, equality of opportunity, equalized odds, and predictive
parity~\citep{hardt2016equality,agarwal2018reductions}, each focusing on equitable
outcomes or error rates across groups. Sources of bias can stem from data (\eg,
under-representation)~\citep{garin2023medical}, algorithms (\eg, prioritizing
accuracy/utility over fairness)~\citep{buolamwini2018gender}, or human involvement
(\eg, subjective labeling)~\citep{zhang2024mitigating}. Perspectives of fairness
include individual fairness (similar individuals receive similar
predictions)~\citep{dwork2012fairness,petersen2021post}, group fairness (equitable
treatment across groups)~\citep{diana2021minimax,chan2024group}, and subgroup fairness
(addressing intersectional identities)~\citep{kuratomi2025subgroup}. Methodologies to
enforce fairness involve pre-processing data (\eg, balancing
representation)~\citep{jang2023difficulty,liu2021just,lahoti2020fairness},
in-processing adjustments (\eg, modifying loss
functions)~\citep{jovanovic2023fare,roy2019mitigating}, and post-processing predictions
(\eg, calibration techniques)~\citep{hardt2016equality,kim2020fact,jang2022group}.
Achieving fairness often involves trade-offs, such as balancing it with
utility~\citep{liu2022accuracy,wang2021understanding} and
interpretability~\citep{jo2023learning}.

Fairness-aware evaluation has an increasing attention in medical imaging tasks, where
model predictions can directly affect downstream diagnostic decisions. Recent studies
have shown that performance can vary across demographic groups due to acquisition
protocols, device related differences, or data imbalance, motivating the inclusion of
fairness metrics alongside performance metrics. For instance, in glaucoma detection, ML
systems have been evaluated not only for accuracy but also for their fairness
performance using equity scaling measurements~\citep{luo2024harvard} or with respect to
demographic features such as gender~\citep{akter2022glaucoma}. Similarly,
multi-objective formalizations have been explored to jointly optimize imaging quality,
diagnostic performance and fairness~\citep{lu2020multiobjective}. However, these works
typically address a single fairness notion and do not provide a general framework
capable of comparing multiple ML systems under several concurrent fairness constraints,
which is the gap our framework aims to bridge.

While several tools and frameworks exist for fairness assessment, their scope differs
significantly from the evaluation problem addressed in this work. For instance,
Fairlearn and its dashboard~\citep{weerts2023fairlearn} and
FACET~\citep{gustafson2023facet} provide analytics for understanding model behavior
across demographic groups, but they evaluate individual model performance rather than
assessing the aggregated structure of a utility-fairness trade-off system achieved with
our framework. This is an important aspect as the assessment of individual models drawn
from a hypothesis class $\mathcal{H}$ may be unstable, whereas evaluating the joint
behavior of an entire set of models provides a more stable, comprehensive and complete
representation of the fairness performance of the ML system in
consideration.~\cite{little2023fairness} proposes a metric that summarizes the
utility-fairness trade-off along a single fairness criterion; however, it does not
generalize to multi-objective scenarios (multiple
utility/fairness).~\cite{liu2022accuracy} model utility-fairness interactions from a
multi-objective perspective, but their formalization focuses on optimization rather
than evaluation, and does not provide model-agnostic tools for analyzing the geometry
or quality of the trade-off optimality. To the best of our knowledge, no existing
framework offers a unified, multi-objective evaluation protocol capable of assessing
multiple utility and fairness criteria jointly, and this aspect motivates the
contribution of our proposed framework.

Whereas ML systems are typically developed (and evaluated) using a single utility
criterion, they are often deployed in scenarios where multiple objectives must be
respected. A modern example of this condition relates to the deployment of ML systems
under one or multiple demographic fairness
constraints~\citep{liu2022accuracy,zhang2021fairer,padh2021addressing}. In this
context, we argue that evaluation techniques cross-pollinated from multi-objective
optimization (MOO) offer a rich set of primitives allowing for a comprehensive
performance characterization under multiple criteria that can streamline system
evaluation in this realm.

The principal aim of MOO is to find solutions that lie on, or are proximate to, the set
of the optimal performance points called the Pareto Front (PF), resulting in a spectrum
of ideal trade-offs among the various objectives. This methodology equips
decision-makers with the means to select the most favorable compromise amidst
conflicting goals, fostering more informed and balanced
decision-making~\citep{wu2001metrics}. The trade-off selection procedure in MOO is
therefore critical and affects the quality of service for the deployed system,
especially in the case of conflicting objectives. Assessing the quality of these
trade-off systems is comparative and encompasses criteria such as proximity to the
Pareto optimal set (convergence), the distribution/spread of the points in the
objective space (diversity), and the cardinality of solutions
(capacity)~\citep{zitzler2003performance}. These criteria are evaluated by MOO specific
performance indicators that have been studied in previous works (refer to
Section~\ref{sec:indicators} for
details)~\citep{tan2002evolutionary,wu2001metrics,van2000measuring,coello2004study}.

Even though the modeling of trade-off for demographic fairness-accuracy PF is
well-known~\citep{wei2022fairness,zietlow2022leveling}, performance indicators for the
quality of the PF have rarely been exploited in the context of
fairness.~\cite{yang2023minimax} developed a bias mitigation framework that
incorporated the Area Under the Curve (AUC) metric, while considering both inter- and
intra-group AUC simultaneously. However, the bias mitigation framework does not provide
an evaluation protocol for the utility-fairness trade-off; instead, it leverages the
AUC to address fairness performance.~\cite{little2023fairness} proposed a scalar
measure of the area under the curve from the trade-off between fairness and accuracy.
The generated curve outlines the empirical Pareto frontier consisting of the highest
attained accuracy within a collection of fitted models at every level of fairness.
Although~\cite{little2023fairness} focuses on similar issues as in this study, it does
not address the challenge of comparing multiple ML systems in high dimensions.
Additionally, the analysis of the PF is superficial, ignoring important performance
indicators for diversity and capacity, providing an incomplete evaluation of compared
ML strategies. To tackle the aforementioned issues, a more flexible evaluation
framework is needed to accommodate different fairness criteria and utility metrics,
facilitating a straightforward comparison and analysis of results from different
algorithms. The method should be model-agnostic, allowing for real-world comparisons
among trade-off systems that may have been optimized using different objectives.
Furthermore, since the utility goals of the model across multiple objectives often
diverge from fairness goals, performance indicators of the optimal PF solutions can
provide a deeper understanding of the trade-offs across these
objectives~\citep{wang2021understanding}. The proposed evaluation framework bridges the
gap between these issues and their solutions, providing a comprehensive guideline on
how this can be achieved in the following sections.

While prior works typically quantify fairness outcomes into a scalar metric or report
the best performing ML system, such simplifications may create ambiguity on the
relationship between different demographic attributes. The evaluation perspective
considered in this study bridges this gap by encouraging analysis across the full
spectrum of utility-fairness trade-offs.
%

\section{Methodology}
\label{sec:method}
In MOO, each individual objective is considered a distinct characteristic that needs to
be optimized to its fullest potential. The trajectory of the optimization is determined
by the objective functions in a cooperative way. Conflicting objectives increase the
complexity of optimization forcing cooperation, thus making it harder to achieve
optimal solutions. This trade-off is measured at evaluation time by using multiple
metrics directly related to each of the objectives.
Likewise, utility and fairness can result in conflicting objectives challenging the
optimization of both through the same ML strategies. A good level of utility is
typically achieved by sacrificing fairness or through less biased models that might
reduce utility.

The proposed evaluation framework considers the trade-off between objectives for
assessment and evaluates the performance of each dimension with a performance metric
tailored for it. There are no limitations on using performance measurements, so any
typically used metric is applicable.

\begin{figure}[t]
    \centering
    \includegraphics[width=0.9\columnwidth]{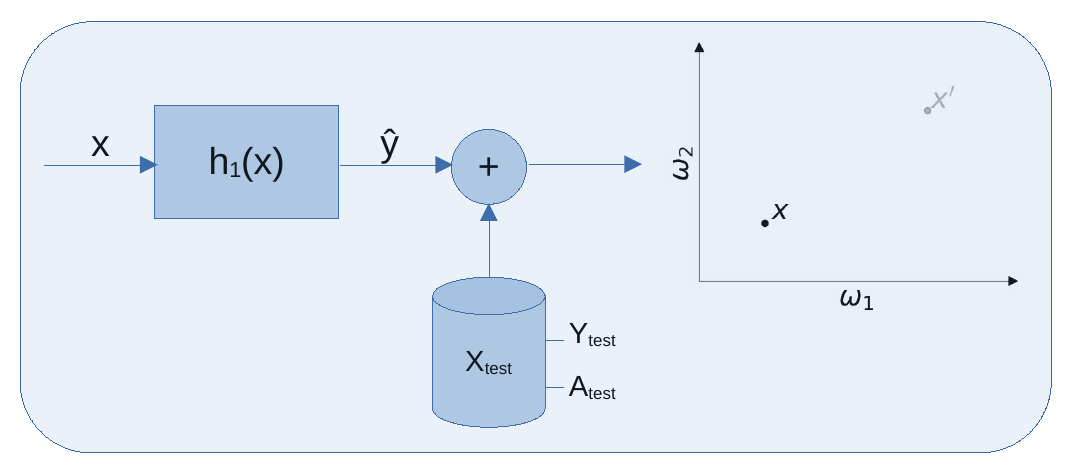}
    \caption{Black-box system evaluation (Scenario 1).}
    \label{fig:scenario_1}
\end{figure}

\subsection{Use Cases}
\label{sec:use-cases}
To formalize the proposed method, we evaluate three ML use cases, which are based on
two scenarios as black-box and white-box, typically found in the literature,
exclusively from a deployment perspective. We explicitly assume that the ML models are
trained and one is only seeking to characterize their performance from a
multi-objective perspective including the model's utility and one to many fairness
objectives.

The first type of scenario considers a ``black-box'' ML system $h_1(x) \in \mathcal{H}$
to provide binary outcomes for an input $x$ such that $\hat{y} = h_1(x)$ where $\hat{y}
    \in \{0,1\}$. To measure the approximate Pareto solution $S$, we assume the
availability of a dataset $X_{test}$ that carries annotations for all considered
objectives, \ie, the expected output of the classification $Y_{test}$, and demographic
attributes $A_{test}$. Fig.~\ref{fig:scenario_1} contains a representation of this
scenario in two optimization dimensions as $\omega_1$ and $\omega_2$. As there is no
tuning possibility to select the model ($\tau=\emptyset$), this test evaluates the
solution in the deployed ML system as it is provided.

\begin{figure}[t]
    \centering
    \includegraphics[width=0.9\columnwidth]{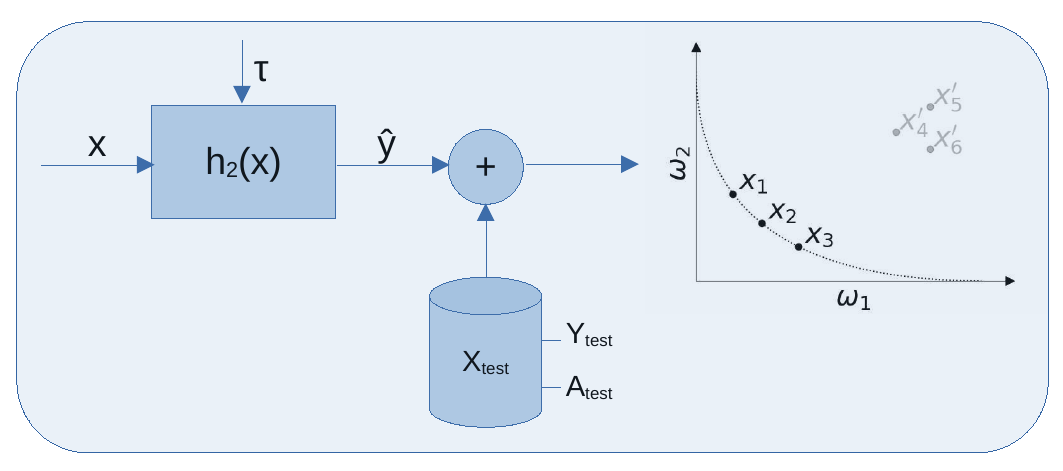}
    \caption{White-box system evaluation (Scenario 2).}
    \label{fig:scenario_2}
\end{figure}

The second scenario defines the evaluation in a ``white-box'' manner for an ML system
$h_2(x) \in \mathcal{H}$ that is tunable over prediction scores (logits) as $\hat{y} =
    h_2(x)$ where $\hat{y} \in [0,1]$. Model selection in $S$ may be achieved via $\tau$ so
that a set of non-dominated solutions filtered by this parameter is available for
performance assessment of the given ML system by using $X_{test}$ alongside $Y_{test}$
and $A_{test}$. This scenario is illustrated in Fig.~\ref{fig:scenario_2} for two
optimization dimensions, $\omega_1$ and $\omega_2$.

In the following, we demonstrate each combination of these two scenarios in three
assessment based use cases. We work with two synthetically generated approximate PF
solutions, \textit{System1} and \textit{System2}, which represent different systems
across the use cases, and evaluate them in a two-dimensional setting to simulate an
assessment in one direction as utility and the other as fairness. Thus, we have an
overall insight into the comparative trade-off performance of different ML systems by
applying possible evaluation strategies commonly encountered during the assessment.

%

\subsubsection{UC-1 - The comparative evaluation of two black-box systems}
In this use case, \textit{System1} and \textit{System2} are considered in a black-box
manner to assess their comparative performance using the proposed evaluation framework.
Since both systems are assumed to be black-box, we only have the model for each as
provided and assess the trade-off performance without any tuning.
\subsubsection{UC-2 - The comparative evaluation of one black-box system with one white-box system as a hybrid case}
This is the use case where the comparative performance of \textit{System1} and
\textit{System2} is evaluated in a hybrid manner by applying black-box and white-box
scenarios. \textit{System1} is assumed to be deployed as it is without any tuning
capability (black-box), and \textit{System2} can be modified to have different settings
based on user preference (white-box).
\subsubsection{UC-3 - The comparative evaluation of two white-box systems}
This use case considers the assessment of \textit{System1} and \textit{System2} as
white-box by tuning them according to specified preferences. Thus, we demonstrate how
the trade-off capacities of the systems can be assessed when tuning is feasible and how
model selection is achieved by fully leveraging their capabilities.

We perform simulations for these use cases in Section~\ref{sec:sim-use-cases} to
exemplify them quantitatively so that it is clarified how the proposed evaluation
framework can be applied for such different assessment strategies given ML systems in
comparison. These simulations are based on the synthetically generated systems,
\textit{System1} and \textit{System2}, and exhibit the PF trend with non-dominated and
dominated points as expected from the utility-fairness trade-off systems.

\subsection{MOO Based Performance Indicators}
\label{sec:framework}
\begin{figure}[t]
    \centering
    \subfloat[Minimization problem\label{fig:obj_min}]{
        \includegraphics[width=0.20\textwidth]{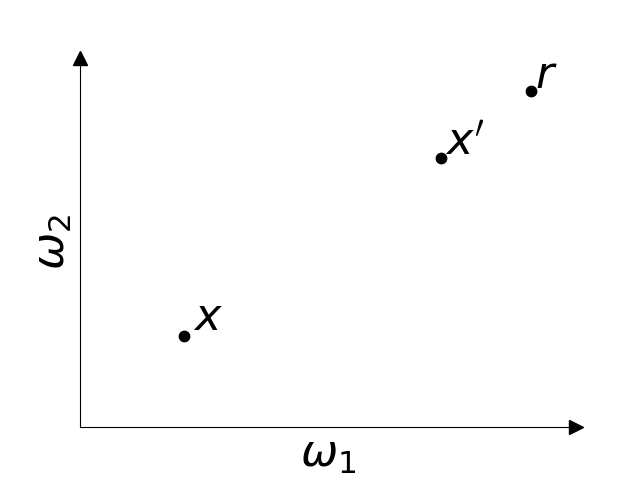}
    }
    \,
    \subfloat[Maximization problem\label{fig:obj_max}]{
        \includegraphics[width=0.20\textwidth]{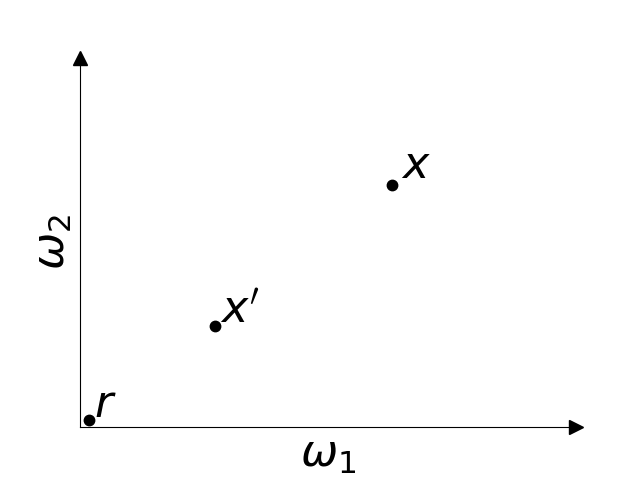}
    }
    \caption{Dominance in bi-objective minimization (a) and maximization (b) problems: $x'$ is dominated by $x$ with respect to the reference point $r$.}
    \label{fig:obj_domin}
\end{figure}

Central to MOO is the concept of the Pareto Front (PF), which delineates the set of all
Pareto optimal solutions. A solution is deemed Pareto optimal if no other solution can
enhance one objective without degrading another. In this regard, solutions residing on
the PF are referred to as non-dominated solutions. More formally, given two points
$\mathbf{x}, \mathbf{x'}$ in the multidimensional solution space $\Omega$ ($\mathbf{x},
    \mathbf{x'} \in \Omega$), the sample $\mathbf{x}$ is said to dominate $\mathbf{x'}$
($\mathbf{x} \prec \mathbf{x'}$), if $\mathbf{x}$ is no worse than $\mathbf{x'}$ in all
considered objective dimensions and is strictly better in at least one of them.
Geometrically, this implies that $\mathbf{x}$ lies farther from the reference point $r
    \in R$, also known as the \textit{nadir} point, which represents the worst possible
outcome in the objective space. In a minimization problem, for example, $\mathbf{x}$
provides a smaller combined value for the target objective compared to $\mathbf{x'}$,
see the illustration in Fig.~\ref{fig:obj_domin}.

%

\begin{figure}[t]
    \centering
    \subfloat[Minimization problem\label{fig:obj_min_pf}]{
        \includegraphics[width=0.20\textwidth]{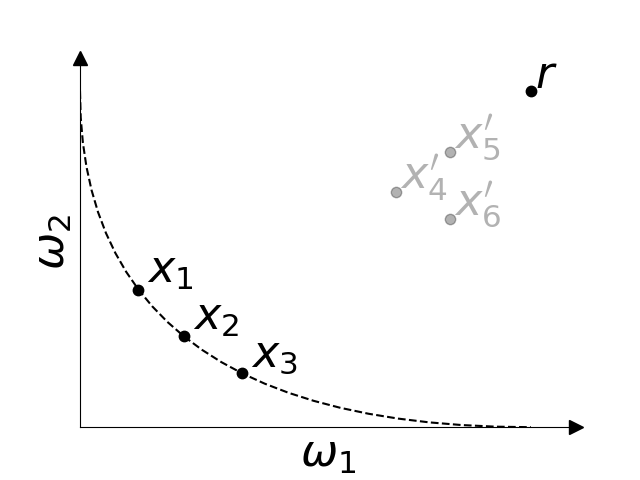}
    }
    \,
    \subfloat[Maximization problem\label{fig:obj_max_pf}]{
        \includegraphics[width=0.20\textwidth]{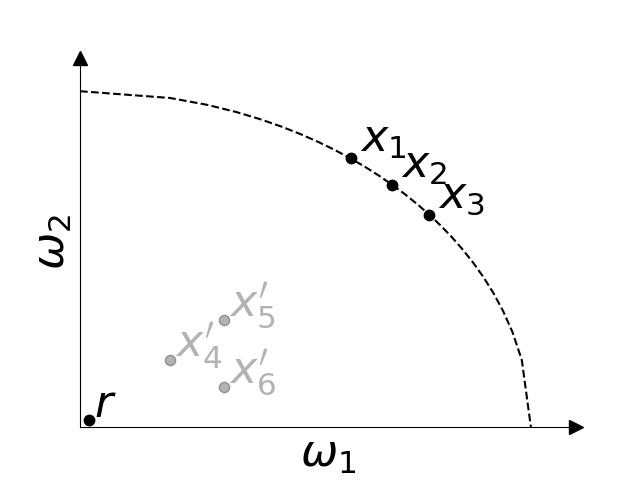}
    }
    \caption{The approximate PF $S$ is shown for both bi-objective minimization (a) and maximization (b) tasks: $x'_i$ is dominated by $x_j$ with respect to the reference point $r$.}
    \label{fig:obj_domin_pf}
\end{figure}

The Pareto optimal set $(\mathcal{P})$ is the set containing all the solutions that are
non-dominated with respect to $\Omega$, defined as:
\begin{equation}
    \label{eq:pareto_optimal}
    \mathcal{P} : = \{\mathbf{x} \in \Omega \;|\; \nexists \mathbf{x'} \in \Omega \quad \mbox{such that} \quad \mathbf{x'} \prec \mathbf{x}  \}
\end{equation}
%
Pareto optimal solutions are called the Pareto set and the image of the Pareto set
constructs the Pareto Front (PF)~\citep{audet2021performance}. We note that, in
real-world problems, the PF is rarely achievable. We refer to suboptimal solutions
approximating the PF as $S$~\citep{zitzler2003performance} as shown in
Fig.~\ref{fig:obj_domin_pf}. We propose expanding on this approach for evaluating ML
systems under multiple fairness constraints. This approach is analogous to the analysis
of Receiver Operating Characteristic (ROC) or detection-error trade-off curves in
classical ML.

Whereas interpretation of a solution set $S$ considering two optimization dimensions
$(\omega_1, \omega_2)$ is straightforward~\citep{little2023fairness}, concurrent
analysis of multiple fairness constraints is typically
done~\citep{buolamwini2018gender,luo2024harvard} as a single degree of fairness by
treating equity performances in isolation from one another. In this type of analysis,
the dependency/correlation between different fairness criteria is not considered and
the evaluation remains oversimplified. However, in a multi-task setting, every
objective may conflict with each other, as one may not be improved without
deteriorating others. This dependency between objectives, as is also the case for
multiple fairness criteria alongside utility, should be projected into one shared space
so that the multiple degrees of evaluation may be achieved in a fused way. To address
this, we propose to characterize the solution set $S$, representing an ML system using
multiple criteria from MOO. These indicators will be assembled in an easy to interpret
table and a plot. Qualitative analysis can still be carried out when the number of
concurrently analyzed objectives is small ($N=2$) or when visual clutter is minimal in
systems with $N>2$. In all cases, the proposed evaluation framework via a PF
characteristic remains usable.

\subsubsection{The Performance Indicators}
\label{sec:indicators}
In the design of metrics for MOO, four complementary performance criteria are typically
considered to analyze the PF optimality: convergence, diversity, convergence-diversity,
and capacity (or cardinality)~\citep{audet2021performance,jiang2014consistencies}.
Measuring strict convergence, which denotes the proximity of the solution to the true
PF, is not often attainable; we therefore focus on the other three properties, and
describe them next.

%

\paragraph{Diversity} This indicator measures how well the non-dominated points are distributed or spread
along the candidate solution set. Uniform Distribution (\textit{UD}) and Overall Pareto
Spread (\textit{OS}) are some diversity measurements based on distribution and spread
characteristics, respectively.

The \textit{UD} indicator~\citep{tan2002evolutionary} evaluates the deviation
characteristic of the distribution for non-dominated solutions, denoted as $X_n$, and
is formulated as:
\begin{equation}
    \label{eq:ud}
    \mathit{UD}(S,\sigma)=\frac{1}{1+D_{\mathit{nc}}(S, \sigma)}
\end{equation}
where
\begin{equation}
    \label{eq:ud_1}
    D_{\mathit{nc}}(S,\sigma)=\sqrt{\frac{1}{|X_n|-1} \sum_{i=1}^{|X_n|} \left(\mathit{nc}(x^i,\sigma)-\mu_{\mathit{nc}(x,\sigma)}\right)^2}
\end{equation}
and
\begin{equation}
    \label{eq:ud_2}
    \mathit{nc}(x^i,\sigma)=|\{x \in X_n \mid \|x-x^i\|<\sigma\}|-1
\end{equation}
$\sigma$ is the niche radius that is problem dependent and can be adjusted based on the
distribution of the candidate solution in the space. $\mu_{\mathit{nc}(x,\sigma)}$ is the mean
of the niche counts, \textit{nc}, and is defined as
\begin{equation}
    \label{eq:ud_3}
    \mu_{\mathit{nc}(x,\sigma)}=\frac{1}{|X_n|} \sum_{j=1}^{|X_n|} \mathit{nc}(x^j,\sigma)
\end{equation}

The \textit{UD} indicator is expected to be higher for a more uniform solution set.
This indicator evaluates how uniform the solution set is spanned in the metric space
based on an upper-bound distance, $\sigma$. For instance, a system with the highest
\textit{UD} value among others exhibits the best performance as its solutions are the
most uniformly distributed. Having a trade-off system with a higher \textit{UD} value
corresponds to a more uniformly spanned set of non-dominated points. This increases the
likelihood of achieving a desired combination of utility with fairness in tuning,
compared to a system with a lower \textit{UD}.
Although the \textit{UD} measures the coverage of the solution space by the candidate
set, it fails to characterize PF as any type of uniformly distributed solution (whether
Pareto optimal or not) may yield high performance in terms of this indicator.

\begin{figure}[t]
    \centering
    \subfloat[Distribution\label{fig:syndata_dist_2d}]{
        \includegraphics[width=0.22\textwidth]{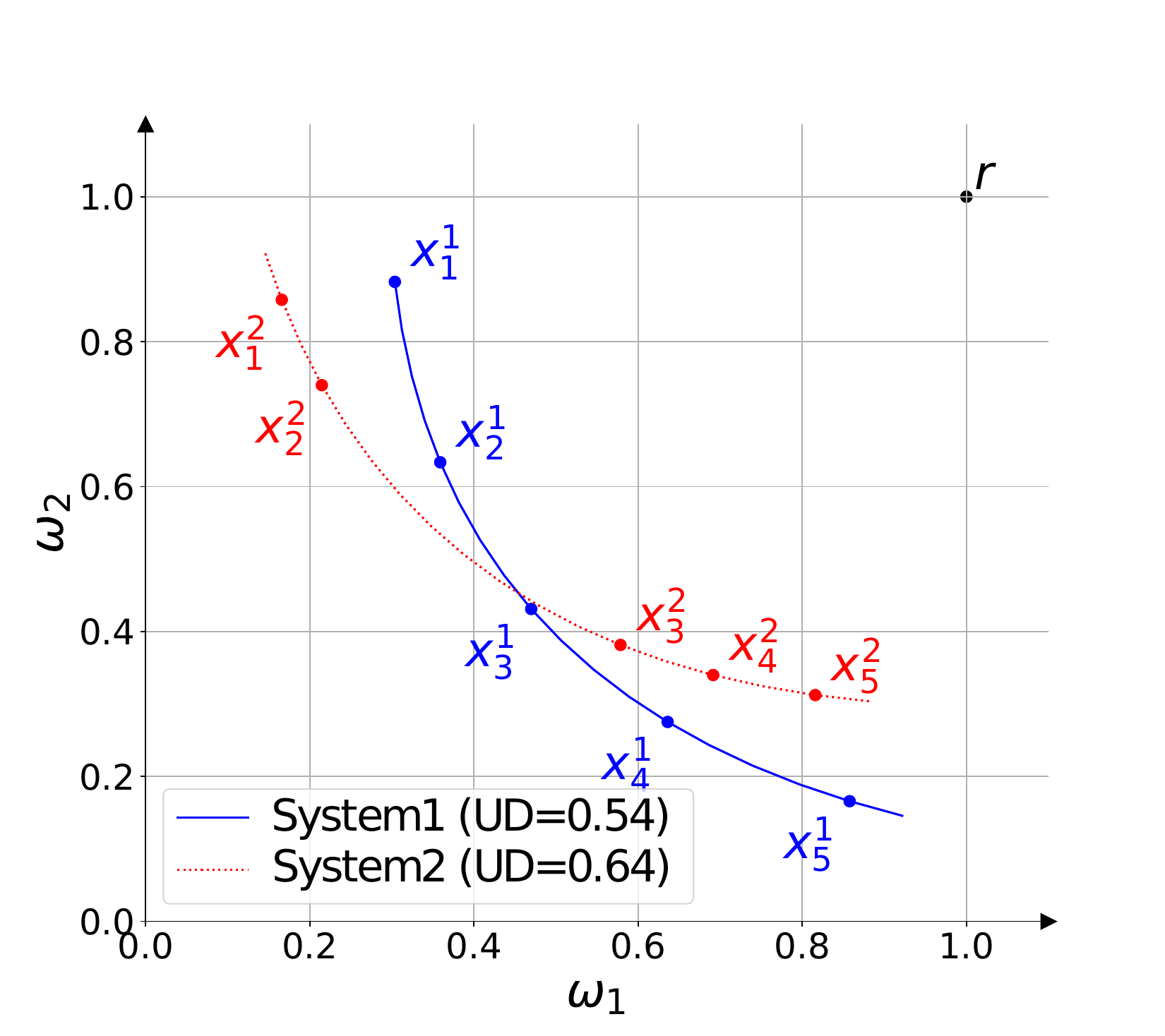}
    }
    \,
    \subfloat[Spread\label{fig:syndata_spread_2d}]{
        \includegraphics[width=0.22\textwidth]{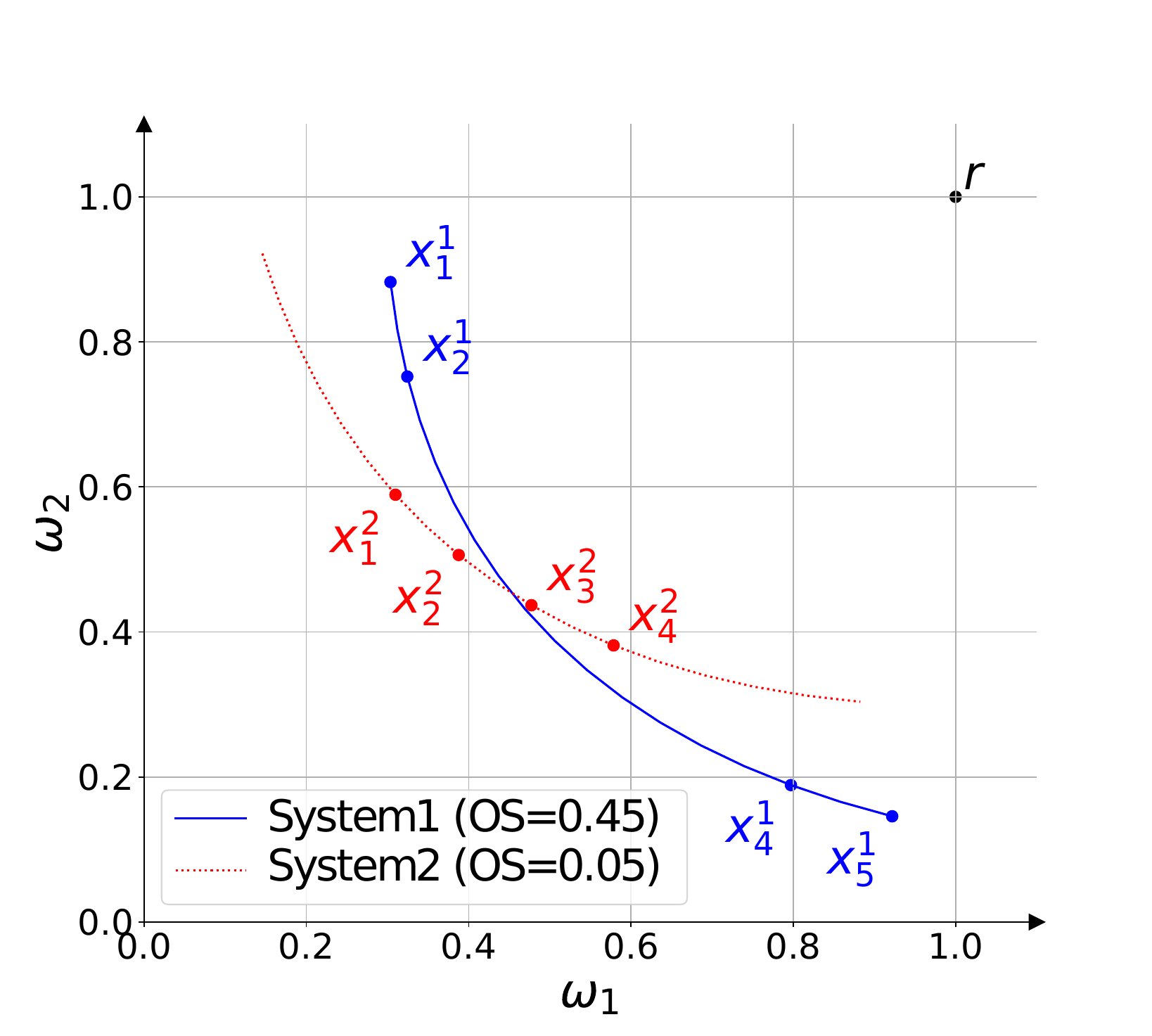}
    }
    \caption{Diversity: (a) \textit{System1} (blue) provides solutions that are less uniformly distributed than \textit{System2} (red) and therefore has lower \textit{UD}.  (b) \textit{System1} (blue) better covers the extremes of the PF approximations and therefore has better spread (larger \textit{OS}) than \textit{System2} (red).}
    \label{fig:syndata_diversity_2d}
\end{figure}

The \textit{OS} indicator~\citep{wu2001metrics} assesses the spread of the solutions
obtained by the trade-off system. For a minimization problem evaluated in $N$ different
dimensions, this indicator is formulated as:
\begin{equation}
    \label{eq:os}
    \mathit{OS}(S,\mathcal{P})=\prod_{i=1}^{N}\left|\frac{\max\limits_{s \in S}s_i-\min\limits_{s \in S}s_i}{\max\limits_{p \in \mathcal{P}}p_{i}-\min\limits_{p \in \mathcal{P}}p_{i}}\right|
\end{equation}
where the nominator and denominator are the absolute difference between the worst and
best points for the candidate solution $S$ and Pareto optimal set $\mathcal{P}$,
respectively. A higher \textit{OS} value indicates a more widely spread solution. This
indicator assesses how well the points from the candidate set spread towards the ideal
of the optimal PF. For instance, a system with a higher \textit{OS} score compared to
others has more points close to the ideal point and fewer ones near the \textit{nadir}
(here, we can access the \textit{nadir} and ideal points without having the exact PF
solution, so there is no requirement to know the PF \textit{a priori}). Having a higher
\textit{OS} value exhibits a more spread characteristic for non-dominated solutions,
leading to an improvement in tuning performance for the trade-off system when the
selection of models around the ideal point is expected. In this study, \textit{OS} is
in the range of $[0,1]$ and there is no transformation applied as it is scaled
compatible with other indicators. Similarly to distribution, this measurement also
fails to analyze Pareto optimality in a comprehensive manner as it only assesses the
extreme cases without considering the entire PF space. In
Fig.~\ref{fig:syndata_diversity_2d}, both \textit{UD} and \textit{OS} indicators are
exemplified in synthetically generated data. \textit{System1} is said to have less
uniformity but more spread than \textit{System2} as its points are more equally
distributed, $\mathit{UD}_{\textit{System1}}=0.54 <
    \mathit{UD}_{\textit{System2}}=0.64$, (Fig.~\ref{fig:syndata_dist_2d}) and closer to
the extreme points, $\mathit{OS}_{\textit{System1}}=0.45 >
    \mathit{OS}_{\textit{System2}}=0.05$, (Fig.~\ref{fig:syndata_spread_2d}).

%
%

During the study, we observed that \textit{OS} can decrease drastically when any
objective fails to cover the full extent of the true PF. This sensitivity arises from
the multiplicative characteristic of \textit{OS} as a small value in one dimension
sharply reduces the final score. To smooth this behavior, we introduce the Average
Spread (\textit{AS}) as a less sensitive variant of \textit{OS}. \textit{AS} simply
replaces the multiplicative operator with summation, and is defined as:
\begin{equation}
    \label{eq:as}
    \mathit{AS}(S,\mathcal{P})=\frac{1}{N}\sum_{i=1}^{N}\left|\frac{\max\limits_{s \in S}s_i-\min\limits_{s \in S}s_i}{\max\limits_{p \in \mathcal{P}}p_{i}-\min\limits_{p \in \mathcal{P}}p_{i}}\right|
\end{equation}
%

\begin{figure}[t]
    \centering
    \includegraphics[width=0.6\columnwidth]{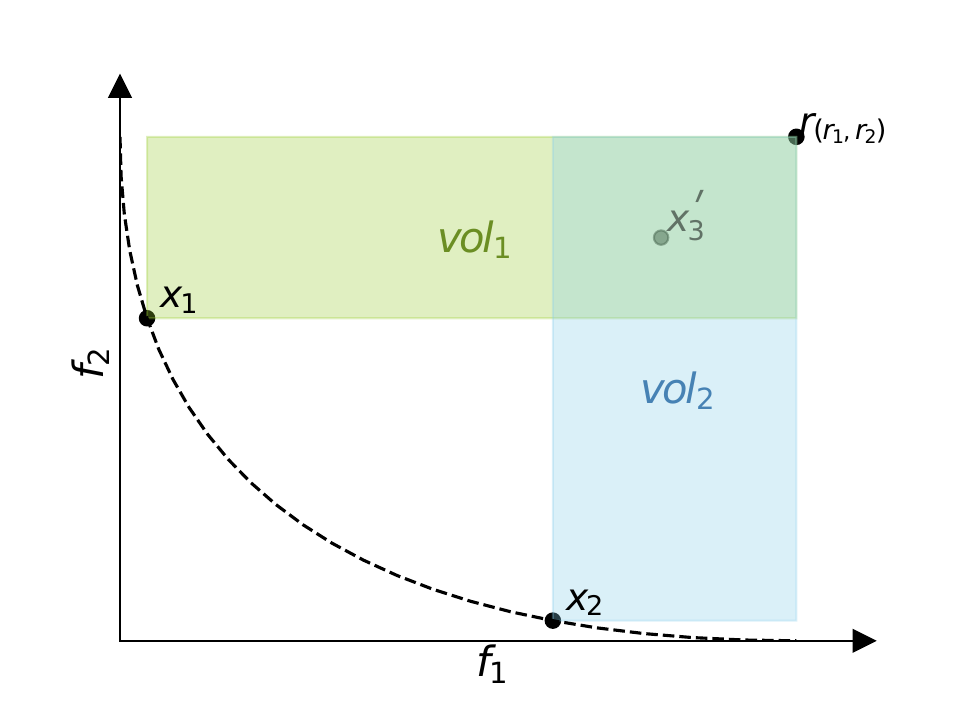}
    \caption{An approximate PF solution with two non-dominated solutions. \textit{HV} is calculated as the union of two volumes associated with these solutions.}
    \label{fig:hv_2d}
\end{figure}

\begin{figure}[t]
    \centering
    \includegraphics[width=0.5\columnwidth]{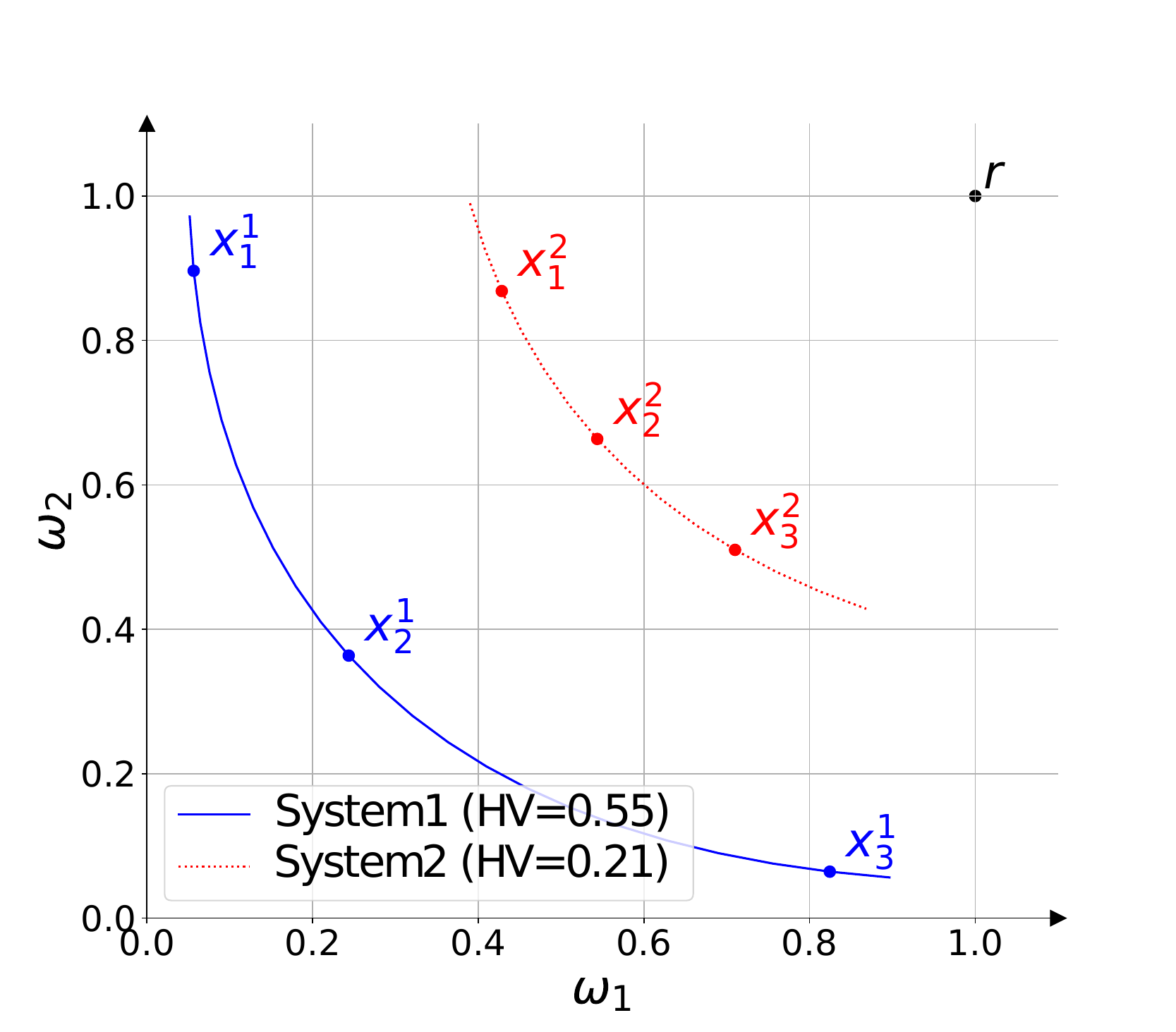}
    \caption{Convergence-Diversity: \textit{System1} (blue) covers a larger volume of the solution space than \textit{System2} (red), relative to the reference $r$ (top-right). Thus, \textit{System1} performs better in terms of \textit{HV}.}
    \label{fig:syndata_hv_2d}
\end{figure}

\paragraph{Convergence-Diversity} This measurement evaluates both convergence and diversity together so that the
proximity (convergence) alongside the distribution/spread (diversity) of the candidate
solution set is projected into a single scalar score. This unary metric measures the
volume of the objective space covered by an approximation set, relying on a reference
point for calculation. The Hypervolume (\textit{HV})~\citep{zitzler1998multiobjective}
takes distribution, spreading, and convergence into account at the same time, making it
unique in this regard. Recognized for its distinctive properties, \textit{HV} is
Pareto-compliant, ensuring that any approximation set achieving maximum quality for a
MOO contains all Pareto optimal solutions. The reference point can be simply attained
by constructing a solution of worst objective function values.
Given a minimization based MOO problem with two objectives, as shown in
Fig.~\ref{fig:hv_2d}, it is expected to have solution sets with points that are in the
best achievable state in the objective space. This should be the case even if the
objectives are conflicting with each other. PF is one possible setting for such cases
with non-dominated solutions. In Fig.~\ref{fig:hv_2d}, $x_1$ and $x_2$ are two
non-dominated solutions drawn from the PF-like solution set (represented as a dashed
curve) with one dominated solution, $x'_3$. The performance of such a solution set may
be evaluated by the \textit{HV} indicator to analyze how optimal the set is in terms of
convergence and diversity. By discarding the dominated solution $x'_3$, which should
not be part of an optimal solution set, the \textit{HV} indicator is calculated as the
union of two volumes constructed between each of the non-dominated solutions, $x_1$ and
$x_2$, and the reference point $r$ that is chosen as one of the poorly performing
solutions in the space. The \textit{HV} formulation is then as follows:
\begin{equation}
    \label{eq:hv_2d}
    \mathit{HV} = \mathit{vol}_{1} \cup \mathit{vol}_{2}=\mathit{VOL}\left(\prod_{i=1}^{2}[x_1^{i},r^{i}] \cup \prod_{i=1}^{2}[x_2^{i},r^{i}]\right)
\end{equation}
The formulation in (\ref{eq:hv_2d}) may be generalized as~\citep{navon2020learning}:
\begin{equation}
    \label{eq:hv}
    \mathit{HV}(S) = \mathit{VOL}\left(\bigcup_{\substack{x \in S \\ x \prec r}} \prod_{i=1}^{N}[x^{i},r^{i}]\right)
\end{equation}
In this study, \textit{HV} is on the scale of $[0,1]$ as every point in the solution
space is represented by measurements between $0$ and $1$. An illustrative example in
Fig.~\ref{fig:syndata_hv_2d} shows how two systems are evaluated in terms of
\textit{HV}. \textit{System1} occupies a larger volume in 2D space compared to
\textit{System2} as it is further away from the reference point. \textit{HV} reflects
this situation with $\mathit{HV}_{\textit{System1}}=0.55 >
    \mathit{HV}_{\textit{System2}}=0.21$.

%

\paragraph{Capacity (or Cardinality)} This measurement quantifies the number of non-dominated points in the candidate
solution set. The Overall Nondominated Vector Generation (\textit{ONVG}) and Overall
Nondominated Vector Generation Ratio (\textit{ONVGR}) are commonly used capacity
indicators. \textit{ONVG}, proposed by~\cite{van2000measuring}, is the number of
non-dominated solutions, $X_n$, in the candidate solution set, $S$, and is formulated
as:
\begin{equation}
    \label{eq:onvg}
    \mathit{ONVG}(S) = |X_n|
\end{equation}
As similarly proposed by~\cite{van2000measuring}, \textit{ONVGR} is the ratio of the
non-dominated solution cardinality to that of $S$, and is defined as:
\begin{equation}
    \label{eq:onvgr}
    \mathit{ONVGR}(S) = \left|\frac{X_n}{S}\right|
\end{equation}

\begin{figure}[t]
    \centering
    \includegraphics[width=0.5\columnwidth]{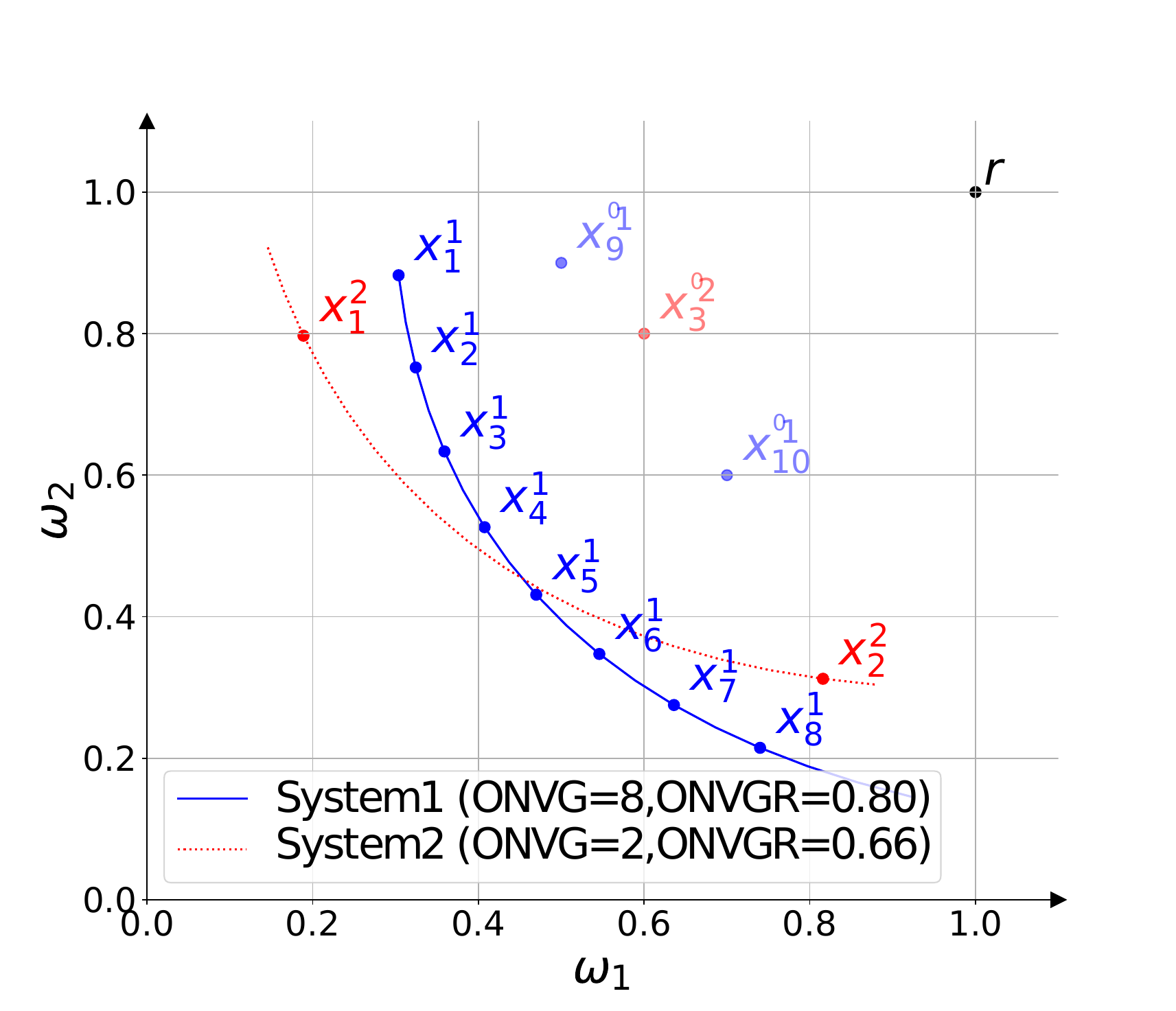}
    \caption{Capacity: \textit{System1} (blue) has more capacity than \textit{System2} (red) as it provides more non-dominated solutions, absolutely (\textit{ONVG}), and relative to the total number of solutions per system (\textit{ONVGR}).}
    \label{fig:syndata_cap_2d}
\end{figure}

Both \textit{ONVG} and \textit{ONVGR} yield higher scores for solution sets with
greater capacity. These capacity-based indicators do not provide an extensive analysis
like convergence or diversity do; they are only used as auxiliary indicators when other
measurements are not discriminative. For instance, we can select a system with a higher
\textit{ONVG} over other systems when the convergence and diversity are the same for
all. Furthermore, they may help analyze the effectiveness of the optimization, as a
higher number of non-dominated points compared to dominated ones is a good indicator of
how well the objective is approximated. Having a larger number of non-dominated points
may also improve tuning the trade-off system as the possibility of finding an expected
combination of utility alongside fairness would increase due to more optimal solutions
in the objective space. We apply a transformation for \textit{ONVG} by normalizing over
the maximum value of it for the systems as
$\widehat{\mathit{ONVG}}=\frac{\mathit{ONVG}}{\max(\mathit{ONVG})}$ to make it in the
same range as others. On the other hand, \textit{ONVGR} is in the range of $[0,1]$ with
$0$ and $1$ indicating the absence of the non-dominated and dominated solutions,
respectively. However, these measurements fail to capture PF optimality as the number
of solutions does not provide information about the Pareto characteristic.
Fig.~\ref{fig:syndata_cap_2d} highlights that \textit{System1} exhibits a more capacity
characteristic compared to \textit{System2} in terms of \textit{ONVG} and
\textit{ONVGR} as it has more non-dominated solutions,
$\mathit{ONVG}_{\textit{System1}}=8 > \mathit{ONVG}_{\textit{System2}}=2$, and a bigger
ratio on overall solutions, $\mathit{ONVGR}_{\textit{System1}}=0.80 >
    \mathit{ONVGR}_{\textit{System2}}=0.66$.

\subsection{Radar Chart: Compact Visualization}
\label{sec:radar}
The assessment of a utility-fairness trade-off system with the aforementioned
performance indicators can be reported as a measurement table. However, it's also
possible to convey this information in different ways such as the illustration in a
chart summarizing all the performance indicators. A radar (spiderweb) chart is such a
compact plot that compares different characteristics in the same projection and allows
for easy comparative analysis of several systems over the same attributes.

%

%
The qualitative analysis resulting from the comparison of utility-fairness trade-offs
with a radar chart makes it possible to select the optimal ML system showing more
capacity, diversity, and convergence-diversity. This overall characteristic of the
systems can also be quantified by calculating the areas occupied by each of them in the
radar chart. The calculation of these areas allows for different systems to be
comparable over very compact quantities compressed by the performance indicators in the
table.

\begin{figure}[t]
    \begin{center}
        \subfloat[Dominance of \textit{System1} (blue) with respect to \textit{System2} (orange) is clearly visible as it occupies a larger volume of the plot. Normalized MOO indicators are used to bind axes to the {[0,1]} scale, improving visual analysis.\label{fig:syndata_radar}]{
            \includegraphics[width=0.80\columnwidth]{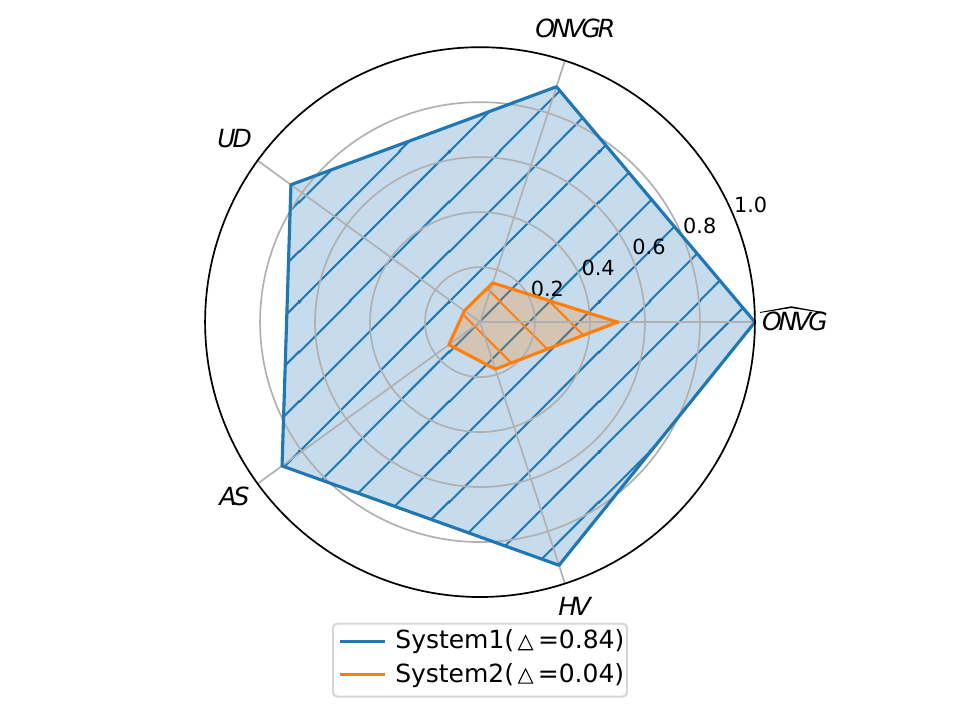}
        }
    \end{center}
    \subfloat[Quantitative results for (a) alongside the inner area of the radar chart.\label{tab:syndata_radar_area}]{
        \resizebox{0.95\columnwidth}{!}{
            \begin{tabular}{l|c|c|c|c|c|c}
                \textbf{ }      & \textbf{Convergence-Diversity} & \multicolumn{2}{c|}{\textbf{Capacity}} & \multicolumn{2}{c|}{\textbf{Diversity}} & \textbf{ }                                                        \\ \hline
                \textbf{ }      & \textbf{ }                     & \multicolumn{2}{c|}{\textbf{ }}        & \textbf{Distribution}                   & \textbf{Spread}    & \textbf{ }                                   \\ \hline
                \textbf{System} & $\bm{\mathit{HV}}$             & $\bm{\widehat{\mathit{ONVG}}}$         & $\bm{\mathit{ONVGR}}$                   & $\bm{\mathit{UD}}$ & $\bm{\mathit{AS}}$ & $\bm{\widehat{\Delta}}$ \\ \hline
                $System 1$      & 0.93                           & 1.00                                   & 0.90                                    & 0.85               & 0.89               & 0.84                    \\
                $System 2$      & 0.18                           & 0.50                                   & 0.15                                    & 0.07               & 0.14               & 0.04                    \\ \hline
            \end{tabular}
        }
    }
    \caption{A sample evaluation with the proposed radar chart. The radar chart provides a summarized assessment of the optimal utility-fairness trade-offs from the evaluated ML systems.}
    \label{fig:syndata_radar_sample}
\end{figure}

We consider the areas in the radar chart as polygons and use the Surveyor's
formula~\citep{braden1986surveyor} for calculation. Given a polygon \textit{Poly} with
$n$ ordered points as counterclockwise in the Cartesian coordinate system, $v_i \in V$,
we define $\mathit{Poly}={v_1,v_2,\ldots,v_n}$ alongside $v_i=(x_i,y_i)$. As we work in
the polar coordinate system to represent the systems in the radar chart, we define
$v_i=(r_i,\Theta_i)$ where $r_i$ is the radius and $\Theta_i$ is the angle. In this
stage, we need to convert the point in polar coordinates to the counterpart in
Cartesian one by $x_i=r_i\cos(\Theta_i)$ and $y_i=r_i\sin(\Theta_i)$. After switching
to the Cartesian coordinate system, we apply the Surveyor's formula as shown below:
\begin{equation}
    \label{eq:aur}
    \Delta_{\mathit{Poly}} = \frac{1}{2}
    \left\{
    \begin{vmatrix}
        x_{1} & x_{2} \\
        y_{1} & y_{2}
    \end{vmatrix}
    +\ldots+
    \begin{vmatrix}
        x_{n-1} & x_{n} \\
        y_{n-1} & y_{n}
    \end{vmatrix}
    +
    \begin{vmatrix}
        x_{n} & x_{1} \\
        y_{n} & y_{1}
    \end{vmatrix}
    \right\}
\end{equation}
where $|.|$ is the $2\times2$ determinant. The calculation of \textit{Area} ($\Delta$)
is then min-max normalized by
$\widehat{\Delta}=\frac{\Delta-\min(\Delta)}{\max(\Delta)-\min(\Delta)}$ to transform
the area range to $[0,1]$. Given a pentagon with 5 dimensions as shown in
Fig.~\ref{fig:syndata_radar}, the theoretical lower and upper bounds of the area are
$0.00$ and $\approx2.37$, respectively. This is analogous to the concept of the Area
Under the Curve (AUC) over Receiver Operating Characteristic (ROC), where an area of
$1.00$ is expected to be the best situation for a system. In the illustrative example
of Fig.~\ref{fig:syndata_radar}, it can be easily seen from the radar chart that
\textit{System1} outperforms \textit{System2} in every dimension. This can be verified
by their respective areas of $0.84$ and $0.04$ (Table~\ref{tab:syndata_radar_area}). We
can also clearly observe that \textit{System1} is closer to the ideal performance than
\textit{System2}, relying only on these areas.

%

\subsection{Deduplication of Solutions}
\label{sec:deduplication}
In our study, each point on the utility-fairness trade-off curve corresponds to an ML
model generated under a specific preference vector. These points form a solution set, $S$,
in a multi-objective space, where the dimensions represent utility and fairness metrics.
However, multiple preference vectors may cause to build ML models that exhibit nearly
identical behavior, and the resulting trade-off system may have (near-)duplicate points
in $S$. Such redundant ML models distort performance indicators by artificially increasing
density or changing the characteristics of the approximated PF.

To alleviate this issue, we use a \textit{deduplication operator}, denoted by
$\mathit{deduplicate}_{\varepsilon}(.)$, which filters out ML models that lie within
$\varepsilon$-neighborhood of each other given a utility-fairness trade-off system. We
leverage the DBSCAN clustering algorithm~\citep{schubert2017dbscan} to retain only the
representative ML models by eliminating redundant ones. Given $S={s_1,s_2,\ldots,s_M}
    \subset \mathcal{R}^N$ where $M$ and $N$ refer to the number of solutions and
objectives, respectively, applying \textit{deduplication operator},
$S'=\mathit{deduplicate}_{\varepsilon}(S)$, eliminates models in similar performance
within $\varepsilon$ formulated as below:
\begin{equation}
    \label{eq:deduplicate}
    \lvert s_i - s_j \rvert > \varepsilon, \quad s_i, s_j \in S'
\end{equation}
We empirically set $\varepsilon=1e-6$ based on experiments conducted on synthetic and
real-world datasets.

\subsection{\textit{A Priori} and \textit{A Posteriori} Analysis} \label{sec:apriori_aposteriori} In ML evaluation,
it is essential to distinguish between \emph{a priori} and \emph{a posteriori}
analysis, as each serves a different role in assessing the generalization and stability
of the systems in comparison. In the context of demographic fairness, we adapt this
concept into our evaluation framework to achieve a similar analysis on utility-fairness
trade-off systems compatible with ML assessment.

In our context, \emph{a priori} analysis refers to identifying Pareto-optimal operating
points by using a validation set, prior to observing the final test data. This
procedure allows the selection of operating combinations of thresholds and sub-ML
models (derived from the trade-off system) that constitute the estimated non-dominated
solution set on the validation set. The goal of this analysis is to perform a real
deployment scenario in which system parameters must be fixed before test-time
evaluation. In contrast, \emph{a posteriori} analysis relies directly on test-set
evaluations and assesses the performance of all sub-ML models without any
pre-selection. While \emph{a posteriori} evaluations provide a complete
characterization of the attainable utility-fairness spectrum, they do not represent a
real generalization scenario, since the selection step is already considered with test
data.

Our framework supports both protocols: \emph{a priori} evaluation performs a realistic
operating mode, whereas \emph{a posteriori} evaluation provides a full diagnostic
understanding of the system's trade-off characteristics. To have \emph{a priori}
evaluation in practice, we allow users to assess two subsets as validation and test by
using the same trained ML model. Based on the validation set, the framework can
determine the threshold/sub-ML combinations that define the Pareto-optimal estimate.
These selections can then be directly applied to the test set to measure how well the
trade-off structure generalizes. In Section~\ref{sec:results}, we provide use cases for
both analyses based on real-world medical problems.

\subsection{Simulations for Use Cases}
\label{sec:sim-use-cases}
\begin{figure}[t]
    \begin{center}
        \subfloat[Dominance of \textit{System1} (blue) with respect to \textit{System2} (orange) is clearly visible from the non-overlapped area.\label{fig:radar_point_point}]{
            \includegraphics[width=0.80\columnwidth]{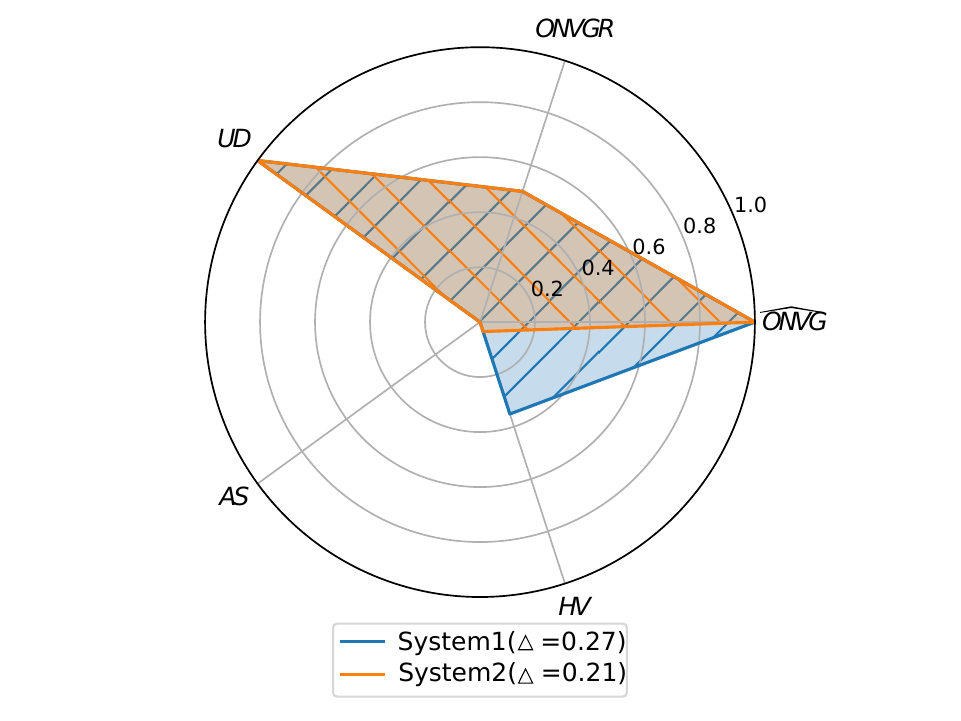}
        }
    \end{center}
    \subfloat[Quantitative results for (a) as black-box.\label{tab:radar_point_point}]{
        \resizebox{0.95\columnwidth}{!}{
            \begin{tabular}{l|c|c|c|c|c|c}
                \textbf{ }       & \textbf{Convergence-Diversity} & \multicolumn{2}{c|}{\textbf{Capacity}} & \multicolumn{2}{c|}{\textbf{Diversity}} & \textbf{ }                                                        \\ \hline
                \textbf{ }       & \textbf{ }                     & \multicolumn{2}{c|}{\textbf{ }}        & \textbf{Distribution}                   & \textbf{Spread}    & \textbf{ }                                   \\ \hline
                \textbf{System}  & $\bm{\mathit{HV}}$             & $\bm{\widehat{\mathit{ONVG}}}$         & $\bm{\mathit{ONVGR}}$                   & $\bm{\mathit{UD}}$ & $\bm{\mathit{AS}}$ & $\bm{\widehat{\Delta}}$ \\ \hline
                \textit{System1} & 0.35                           & 1.00                                   & 0.50                                    & 1.00               & 0.00               & 0.27                    \\
                \textit{System2} & 0.04                           & 1.00                                   & 0.50                                    & 1.00               & 0.00               & 0.21                    \\ \hline
            \end{tabular}
        }
    }
    \caption{Simulation for UC-1.}
    \label{fig:syndata_point_point}
\end{figure}

The first use case, UC-1, focuses on black-box testing of \textit{System1} and
\textit{System2}, corresponding to the first scenario (Fig.~\ref{fig:scenario_1}). As
both systems have just $1$ non-dominated solution, they have the same
$\widehat{\mathit{ONVG}}$ and \textit{ONVGR} values of $1.00$ and $0.50$ respectively.
In terms of the diversity measurements, it is not possible to evaluate both systems as
there exists only $1$ non-dominated solution. \textit{System1} has a higher \textit{HV}
score than \textit{System2} because its non-dominated solution is farther from the
\textit{nadir} point than the non-dominated solution of the other, resulting in a
larger volume. Table~\ref{tab:radar_point_point} and Fig.~\ref{fig:radar_point_point}
show this comparison quantitatively and qualitatively. The radar chart in
Fig.~\ref{fig:radar_point_point} illustrates the performance gap and \textit{HV}
dominance of \textit{System1} over \textit{System2}. The difference in the area is
$\widehat{\Delta}=\widehat{\Delta_1}-\widehat{\Delta_2}=0.27-0.21=0.06$ and it arises
solely from the \textit{HV} differences between the two systems. In this case, the
proposed evaluation framework simplifies the comparison using a single indicator,
\textit{HV}, as the other indicators do not play a role in the model selection
decision.

%

\begin{figure}[t]
    \begin{center}
        \subfloat[Dominance of \textit{System2} (orange) with respect to \textit{System1} (blue) is visible as it occupies a larger volume of the plot.\label{fig:radar_line_point}]{
            \includegraphics[width=0.80\columnwidth]{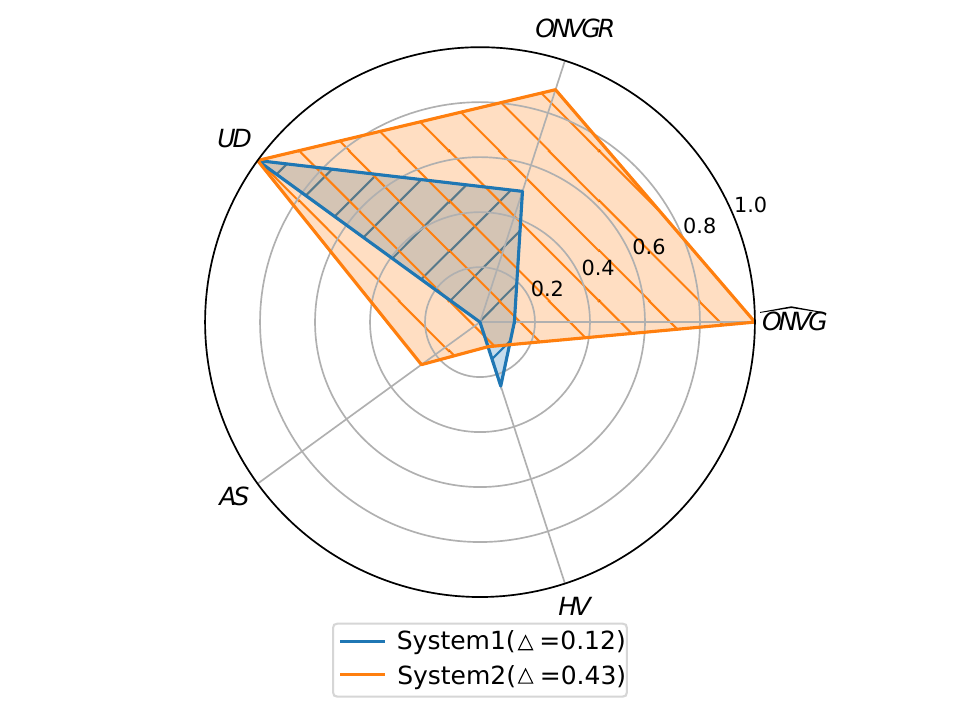}
        }
    \end{center}
    \subfloat[Quantitative results for (a) as black- and white-box.\label{tab:radar_line_point}]{
        \resizebox{0.95\columnwidth}{!}{
            \begin{tabular}{l|c|c|c|c|c|c}
                \textbf{ }       & \textbf{Convergence-Diversity} & \multicolumn{2}{c|}{\textbf{Capacity}} & \multicolumn{2}{c|}{\textbf{Diversity}} & \textbf{ }                                                        \\ \hline
                \textbf{ }       & \textbf{ }                     & \multicolumn{2}{c|}{\textbf{ }}        & \textbf{Distribution}                   & \textbf{Spread}    & \textbf{ }                                   \\ \hline
                \textbf{System}  & $\bm{\mathit{HV}}$             & $\bm{\widehat{\mathit{ONVG}}}$         & $\bm{\mathit{ONVGR}}$                   & $\bm{\mathit{UD}}$ & $\bm{\mathit{AS}}$ & $\bm{\widehat{\Delta}}$ \\ \hline
                \textit{System1} & 0.24                           & 0.12                                   & 0.50                                    & 1.00               & 0.00               & 0.12                    \\
                \textit{System2} & 0.09                           & 1.00                                   & 0.89                                    & 1.00               & 0.26               & 0.43                    \\ \hline
            \end{tabular}
        }
    }
    \caption{Simulation for UC-2.}
    \label{fig:syndata_line_point}
\end{figure}

In the second use case, UC-2, we perform hybrid testing with black- (\textit{System1},
Fig.~\ref{fig:scenario_1}) and white-box (\textit{System2}, Fig.~\ref{fig:scenario_2})
cases. $8$ different non-dominated solutions are considered out of $25$, adjusting
$\tau$ for \textit{System2} against a single non-dominated solution for
\textit{System1}, as it is not tunable. Table~\ref{tab:radar_line_point} contains the
indicator scores for both systems. \textit{System2} outperforms \textit{System1} for
$\widehat{\mathit{ONVG}}$ as it has more non-dominated solutions but results in a lower
\textit{HV} with a score of $0.09$. The distribution indicator is not informative as
\textit{UD} is same for both systems, and the spread is better for \textit{System2}
with an \textit{AS} of $0.26$ compared to $0.00$ for \textit{System1}. This can be
observed in the radar chart shown in Fig.~\ref{fig:radar_line_point}. We can end up
with a decision that \textit{System2} outperforms \textit{System1} overall, as seen in
the difference of the areas
$\widehat{\Delta}=\widehat{\Delta_2}-\widehat{\Delta_1}=0.43-0.12=0.31$.

%

\begin{figure}[t]
    \begin{center}
        \subfloat[Dominance of \textit{System1} (blue) with respect to \textit{System2} (orange) is clearly visible from the non-overlapped area as it occupies
            a larger volume of the plot.\label{fig:radar_line_line}]{
            \includegraphics[width=0.80\columnwidth]{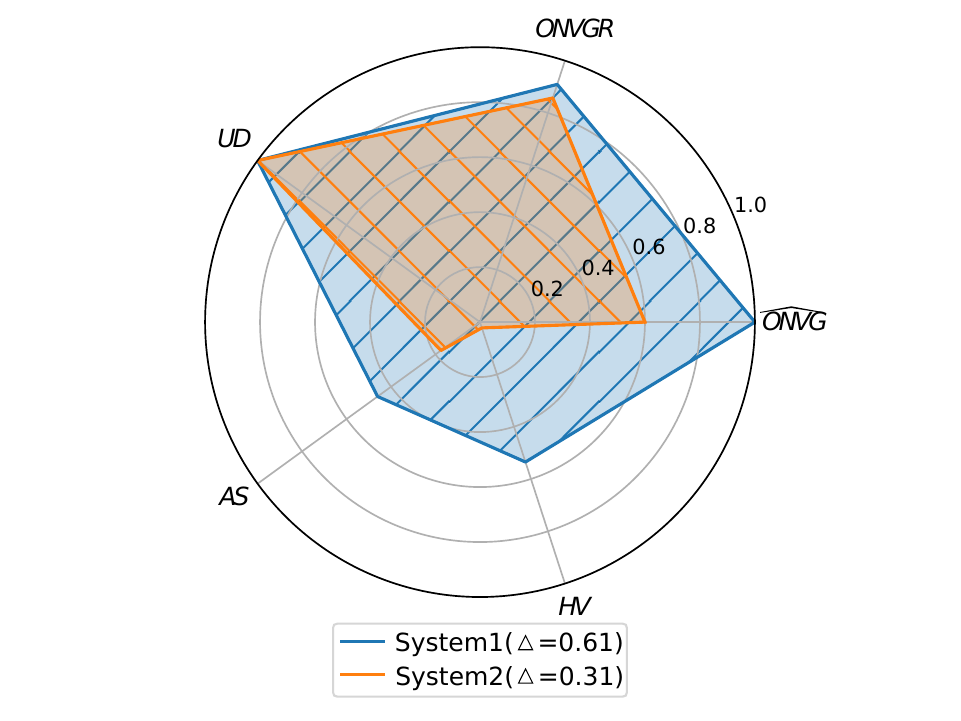}
        }
    \end{center}
    \subfloat[Quantitative results for (a) as white-box.\label{tab:radar_line_line}]{
        \resizebox{0.95\columnwidth}{!}{
            \begin{tabular}{l|c|c|c|c|c|c}
                \textbf{ }       & \textbf{Convergence-Diversity} & \multicolumn{2}{c|}{\textbf{Capacity}} & \multicolumn{2}{c|}{\textbf{Diversity}} & \textbf{ }                                                        \\ \hline
                \textbf{ }       & \textbf{ }                     & \multicolumn{2}{c|}{\textbf{ }}        & \textbf{Distribution}                   & \textbf{Spread}    & \textbf{ }                                   \\ \hline
                \textbf{System}  & $\bm{\mathit{HV}}$             & $\bm{\widehat{\mathit{ONVG}}}$         & $\bm{\mathit{ONVGR}}$                   & $\bm{\mathit{UD}}$ & $\bm{\mathit{AS}}$ & $\bm{\widehat{\Delta}}$ \\ \hline
                \textit{System1} & 0.54                           & 1.00                                   & 0.91                                    & 1.00               & 0.46               & 0.61                    \\
                \textit{System2} & 0.02                           & 0.60                                   & 0.86                                    & 1.00               & 0.18               & 0.31                    \\ \hline
            \end{tabular}
        }
    }
    \caption{Simulation for UC-3.}
    \label{fig:syndata_line_line}
\end{figure}

The last use case, UC-3, considers white-box testing for both \textit{System1} and
\textit{System2}, matching the second scenario (Fig.~\ref{fig:scenario_2}). $10$ and
$6$ different non-dominated solutions were considered out of $25$ for \textit{System1}
and \textit{System2}, respectively. As seen in Table~\ref{tab:radar_line_line},
\textit{System1} outperforms \textit{System2} in terms of $\widehat{\mathit{ONVG}}$,
\textit{ONVGR}, \textit{HV}, and \textit{AS} with same performance in terms of
\textit{UD}. This can also be interpreted through a visual inspection of the radar
chart in Fig.~\ref{fig:radar_line_line}. Finally, the areas of both systems confirm
this as well, $\widehat{\Delta}=\widehat{\Delta_1}-\widehat{\Delta_2}=0.61-0.31=0.30$.

We can derive some conclusions about performance indicators based on the observations
regarding the use cases above. Firstly, the black-box case only gives a partial
characterization of the assessment since the ML systems are not tunable. In this case,
we obtain non-discriminative values for $\widehat{\mathit{ONVG}}$, \textit{ONVGR},
\textit{UD}, and \textit{AS} as seen in Fig.~\ref{fig:syndata_point_point}. Secondly,
diversity is another issue when working with a low number of solutions. A system with
only $1$ solution does not allow to evaluate \textit{AS} as there must be at least two
solutions to measure the distance between the extreme points from the system and PF. A
single point solution generates an \textit{AS} score of $0.00$ as seen in
\textit{System1} shown in Table~\ref{tab:radar_line_point}. Thirdly, \textit{HV} may be
the first decision point to select the system, which exhibits more PF characteristics,
as it covers every aspect of convergence, diversity and capacity. There may be opposite
cases between \textit{HV} and the other indicators as seen in
Fig.~\ref{fig:syndata_line_point}. A good indication of $\widehat{\mathit{ONVG}}$,
\textit{ONVGR}, \textit{UD}, and \textit{AS} does not work as well as evaluating the
performance over \textit{HV}, which has been proven to be a more reliable measurement
for PF~\citep{audet2021performance,jiang2014consistencies}.

\section{Empirical Validation}
\label{sec:empirical}
In this section, we demonstrate the effectiveness of the proposed evaluation framework
empirically using three fairness-aware medical imaging datasets.

\subsection{Dataset Description}
\label{sec:dataset}
To empirically validate the proposed evaluation framework, we use three medical
imaging datasets with demographic attributes: the Harvard Glaucoma Fairness (HGF)
dataset~\citep{luo2024harvard}, the Shenzhen Chest X-ray dataset~\citep{jaeger2014two},
and the mBRSET retinal dataset~\citep{wu2025portable}. These datasets were selected
because they jointly capture demographic and clinical diversity, and enable the evaluation
of utility-fairness trade-offs under both modality-based and population-based conditions.

\paragraph{Harvard Glaucoma Fairness (HGF).}
The HGF dataset includes cases with a retinal nerve disease called glaucoma as well as
samples of healthy patients. Glaucoma is twice as common in Black patients compared to
other races, and more prevalent in men~\citep{khachatryan2019primary,luo2024harvard}.
The dataset comprises $3300$ two-dimensional retinal nerve fiber layer thickness
(RNFLT) maps from three racial groups, namely Asian, Black, and White, with a
resolution of $200 \times 200$ pixels. Color intensity represents retinal nerve fiber
thickness in micrometers around the optic disc. The glaucoma/non-glaucoma ratio in the
dataset is $53.0\%/47.0\%$. The prevalence of glaucoma in Asians, Blacks, and Whites in
the HGF dataset is equally balanced at $33.3\%$, and the gender distribution is as
$54.9\%$ and $45.1\%$ for females and males, respectively. The ophthalmologic images
from HGF are linked to sensitive attributes for race, gender, age, and ethnicity. This
dataset provides a clear setting for assessing demographic disparities, as the higher
glaucoma prevalence among Black patients contradicts the relative scarcity of such
samples in ophthalmology data. This fact creates a fairness challenge that represents
real-world imbalance in healthcare AI systems and provides a solid use case for our
framework.

\paragraph{Shenzhen Chest X-ray Dataset.}
The Shenzhen dataset is a public chest X-ray collection designed for computer-aided
screening of pulmonary diseases, particularly tuberculosis (TB). The dataset contains
$662$ frontal chest X-rays with a distribution of normal and TB-positive cases as $326$
and $336$, respectively. Image resolutions vary with an approximation of $3000 \times
    3000$ pixels. The dataset has sensitive attributes for age and gender, providing a
complementary use case for evaluating fairness across clinical subgroups with chest
radiographs.

\paragraph{mBRSET Retinal Dataset.}
The mBRSET dataset is a retinal imaging resource and designed to enable the
benchmarking of automated ophthalmologic screening models under different demographic
conditions. The dataset consists of $5164$ retinal images from $1291$ subjects,
including fundus photography modality. Each sample is annotated for ocular diseases
such as diabetic retinopathy and macular edema, alongside demographic attributes such
as patient age, gender, and obesity. The image resolution varies in height and width,
ranging from $874$ and $951$ to $2304$ to $2984$ pixels, respectively. The dataset
exhibits a gender imbalance, with female patients compared to male ones as
$65.1\%/34.9\%$. The patient age has an average of $61.4$ years with a standard
deviation of $11.6$.

\begin{figure}[t]
    \centering
    \subfloat[HGF RNFLT map.\label{fig:hgf_example}]
    {
        \includegraphics[width=0.13\textwidth]{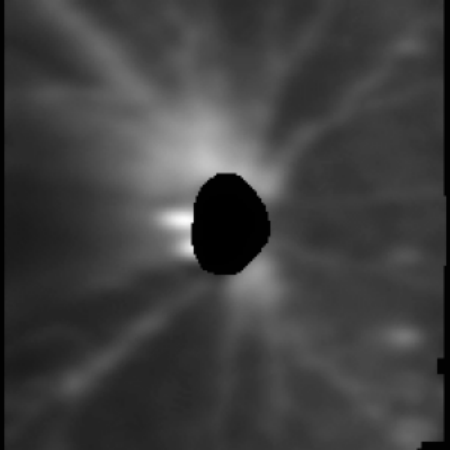}
    }
    \hspace{0.3cm}
    \subfloat[Shenzhen X-ray.\label{fig:shenzhen_example}]
    {
        \includegraphics[width=0.13\textwidth]{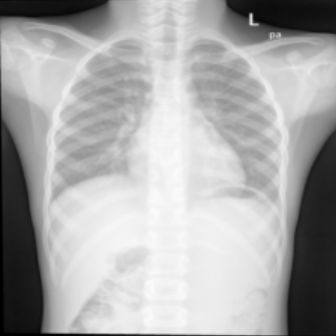}
    }
    \hspace{0.3cm}
    \subfloat[mBRSET fundus.\label{fig:mbrset_example}]
    {
        \includegraphics[width=0.13\textwidth]{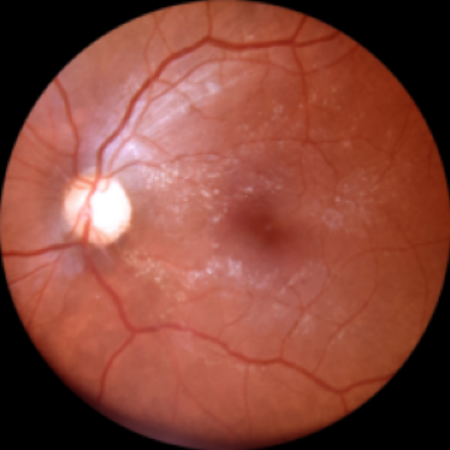}
    }
    \caption{Sample images from the datasets used in the empirical study: (a) RNFLT map
        from HGF for glaucoma detection; (b) chest X-ray from Shenzhen for pulmonary disease
        screening; and (c) retinal fundus image from mBRSET for diabetic retinopathy classification.}
    \label{fig:dataset_examples}
\end{figure}

By jointly employing these three datasets, our empirical evaluation covers
complementary fairness perspectives across medical imaging. HGF provides a clear
setting for analyzing demographic and disease prevalence imbalances in glaucoma with
race and gender labels. The Shenzhen dataset extends this perspective by providing
chest X-ray imagery annotated with sensitive attributes such as gender, enabling
fairness assessment in a different clinical modality. Finally, mBRSET provides clinical
diversity across ocular conditions by having a broader diagnostic context with
different retinal diseases. Together, these datasets enable a more comprehensive
examination of how the proposed framework evaluates utility-fairness trade-offs across
different medical imaging modalities, disease types, and demographic distributions.

\subsection{System Definitions and Configuration}
\label{sec:configuration}
To properly demonstrate the capabilities of the proposed evaluation framework,
we formulate the utility-fairness trade-off as a bi-objective optimization problem
through two distinct systems denoted as \textit{System1} and \textit{System2}, which
differ from the systems considered in previous experiments.
For the HGF dataset, both systems are developed using Pareto HyperNetworks (PHNs),
which are based on hypernetworks~\citep{ha2016hypernetworks}, and are designed to learn
Pareto-optimal solutions across multiple objectives~\citep{navon2020learning}. The
PHN formulation allows the generation of sub-neural networks (sub-NNs) that represent
different levels of compromise between diagnostic utility and fairness, enabling continuous
exploration along the PF. In particular, each sub-NN corresponds to a specific
utility-fairness trade-off level, \ie, one sub-model represents a diagnostic
performance with higher utility but lower fairness, whereas another has higher equity
at the cost of classification accuracy. Each objective combination is
represented by a preference vector that encodes the relative weighting between the
two objectives. In the HGF setting, all sub-NNs are evaluated directly on the
test set, and the full spectrum of utility-fairness outcomes is characterized as
\emph{a posteriori}, without a validation-driven selection of operating points.

Each PHN employs a ResNet-18 backbone as a shared encoder that generates parameters for
sub-NNs corresponding to different preference vectors. A total of $25$ preference
vectors are uniformly sampled from a Dirichlet distribution with $\alpha=0.2$ to
represent distinct utility-fairness trade-offs. The utility objective is defined by
Binary Cross-Entropy (BCE) loss, and the fairness objective is modeled using a
differentiable relaxation over Equalized Odds (EO) criterion, which captures
disparities in true positive and false positive rates (TPR, FPR) across demographic
subgroups. The combined objective is jointly optimized through back-propagation to
dynamically generate Pareto-optimal solutions that balance diagnostic utility and
fairness.

For the mBRSET and Shenzhen datasets, \textit{System1} and \textit{System2} use two
different model architectures. \textit{System1} is based on the DenseNet
topology~\citep{huang2017densely}, and this setting follows a training mechanism
performed directly under a bi-objective loss combining BCE for utility and EO-based
fairness penalties to represent an independent ML model that corresponds to a specific
preference. Unlike the joint optimization characteristic of the PHN, here, each
utility-fairness trade-off is modeled as a separate ML model. \textit{System2} employs
a LoRA-enabled ViT-Small model~\citep{hu2022lora,dosovitskiy2020image}, where low-rank
adaptation layers are inserted into the self-attention blocks to enable lightweight
fairness-aware fine-tuning. For both datasets, the hyperparameters (\eg, learning rate,
optimizer, and batch size) follow a fixed configuration within each system. For the
mBRSET, PF operating points are selected on a validation set and then evaluated on the
test set, following \emph{a priori} analysis of the utility-fairness trade-offs,
whereas for Shenzhen, all operating points are evaluated directly on the validation
set, corresponding to an \emph{a posteriori} analysis.

\textbf{\textit{System1}} serves as a baseline fairness configuration across all datasets.
It focuses on optimizing fairness with respect to a \emph{binary sensitive attribute}
in each case. For the HGF dataset, this system minimizes EO disparities between \emph{male}
and \emph{female} patients, targeting \emph{gender fairness}. For the mBRSET, the system
similarly minimizes EO disparities between obese and non-obese patients, capturing
\emph{obesity-related fairness}. For the Shenzhen dataset, the system again focuses on
\emph{gender fairness} by minimizing EO disparities between same subgroups.

\textbf{\textit{System2}} extends this baseline to investigate dataset specific
demographic variation. In the HGF setting, this system  shifts the fairness
objective from gender to race by minimizing EO disparities between \emph{Asian}, \emph{Black},
and \emph{White} subgroups. For the mBRSET dataset, the LoRA-enabled ViT-Small model
uses the same EO-based \emph{obesity fairness} objective. Similarly, for the Shenzhen
dataset, the same topology minimizes EO disparities between \emph{male} and \emph{female}
patients.

All systems within the same dataset share identical training procedures and
hyperparameters to ensure comparability. For PHN-based systems, optimization is
performed using Adam optimizer with a learning rate of $1e-4$ and batch size of $256$.
DenseNet- and LoRA-ViT-Small-based systems employ dataset-specific but fixed
hyperparameters across \textit{System1} and \textit{System2} using Adam optimizer with
a batch size of $16$. However, learning rates differ as $5e-5$ and $1e-4$ for the
mBRSET and Shenzhen datasets, respectively. This shared setup ensures that differences
in observed performance arise solely from the fairness objective being optimized to
provide a controlled analysis of utility-fairness trade-offs.


\subsection{Evaluation Results}
\label{sec:results}
\begin{figure}[t]
    \begin{center}
        \subfloat[Pareto plot.\label{fig:pareto_mbrset_obesity}]{
            \includegraphics[width=0.80\columnwidth]{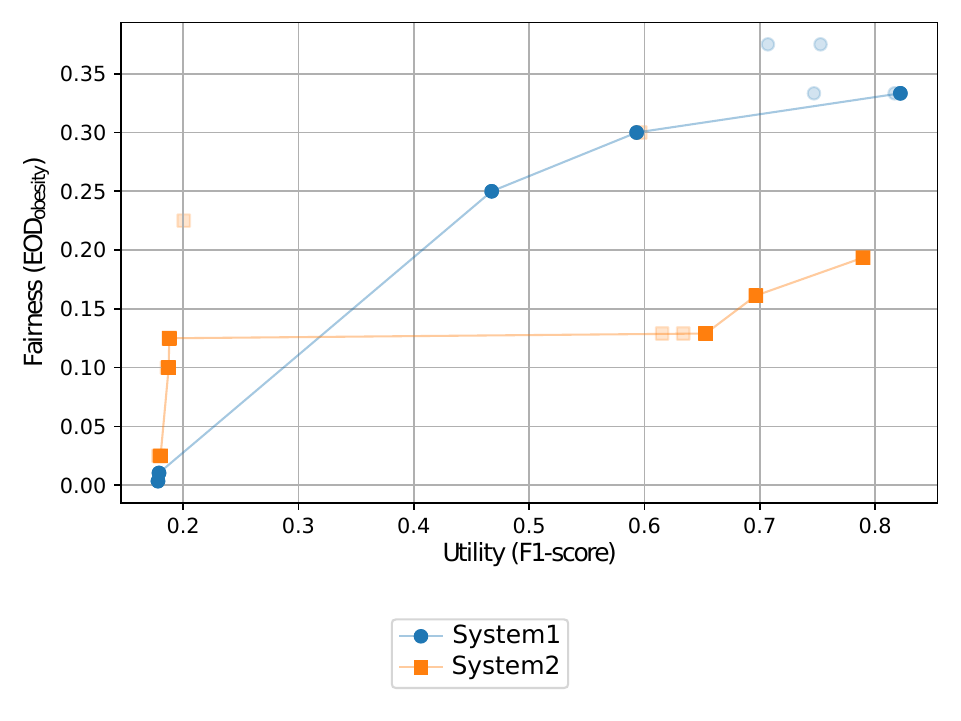}
        }
        \hspace{0.3cm}
        \subfloat[Radar chart.\label{fig:radar_mbrset_obesity}]{
            \includegraphics[width=0.80\columnwidth]{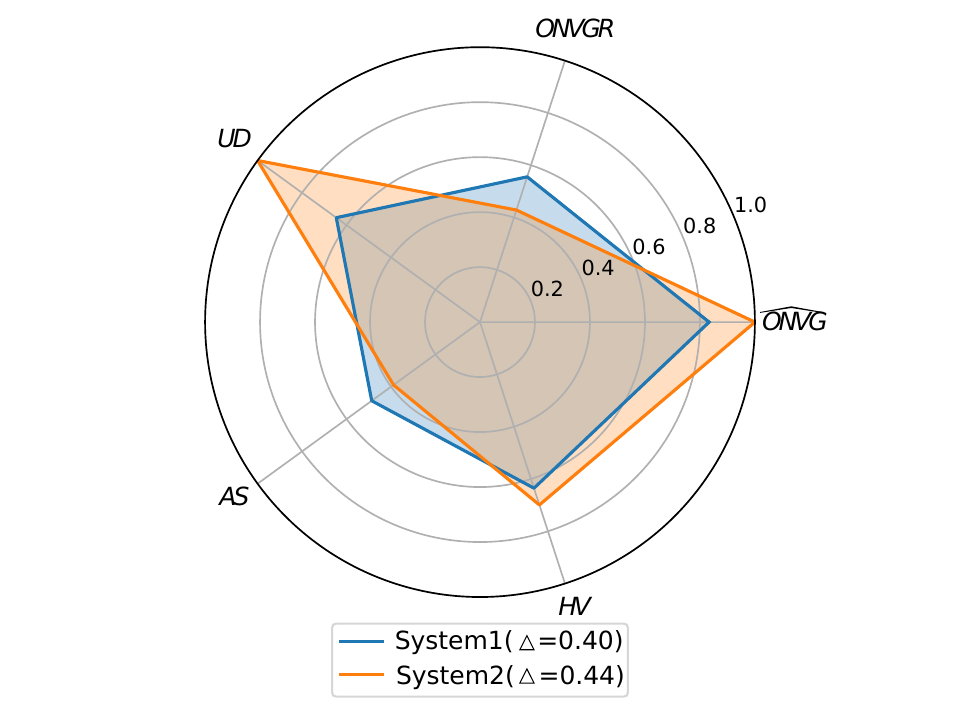}
        }
    \end{center}
    \subfloat[Quantitative results.\label{tab:radar_mbrset_obesity}]{
        \resizebox{0.95\columnwidth}{!}{
            \begin{tabular}{l|c|c|c|c|c|c}
                \textbf{ }       & \textbf{Convergence-Diversity} & \multicolumn{2}{c|}{\textbf{Capacity}} & \multicolumn{2}{c|}{\textbf{Diversity}} & \textbf{ }                                                        \\ \hline
                \textbf{ }       & \textbf{ }                     & \multicolumn{2}{c|}{\textbf{ }}        & \textbf{Distribution}                   & \textbf{Spread}    & \textbf{ }                                   \\ \hline
                \textbf{System}  & $\bm{\mathit{HV}}$             & $\bm{\widehat{\mathit{ONVG}}}$         & $\bm{\mathit{ONVGR}}$                   & $\bm{\mathit{UD}}$ & $\bm{\mathit{AS}}$ & $\bm{\widehat{\Delta}}$ \\ \hline
                \textit{System1} & 0.64                           & 0.83                                   & 0.56                                    & 0.65               & 0.49               & 0.40                    \\
                \textit{System2} & 0.70                           & 1.00                                   & 0.43                                    & 1.00               & 0.39               & 0.44                    \\ \hline
            \end{tabular}
        }
    }
    \caption{DenseNet/LoRA-ViT-Small on mBRSET.}
    \label{fig:mbrset_obesity}
\end{figure}
\begin{figure}[t]
    \begin{center}
        \subfloat[Pareto plot.\label{fig:pareto_shenzhen_gender}]{
            \includegraphics[width=0.80\columnwidth]{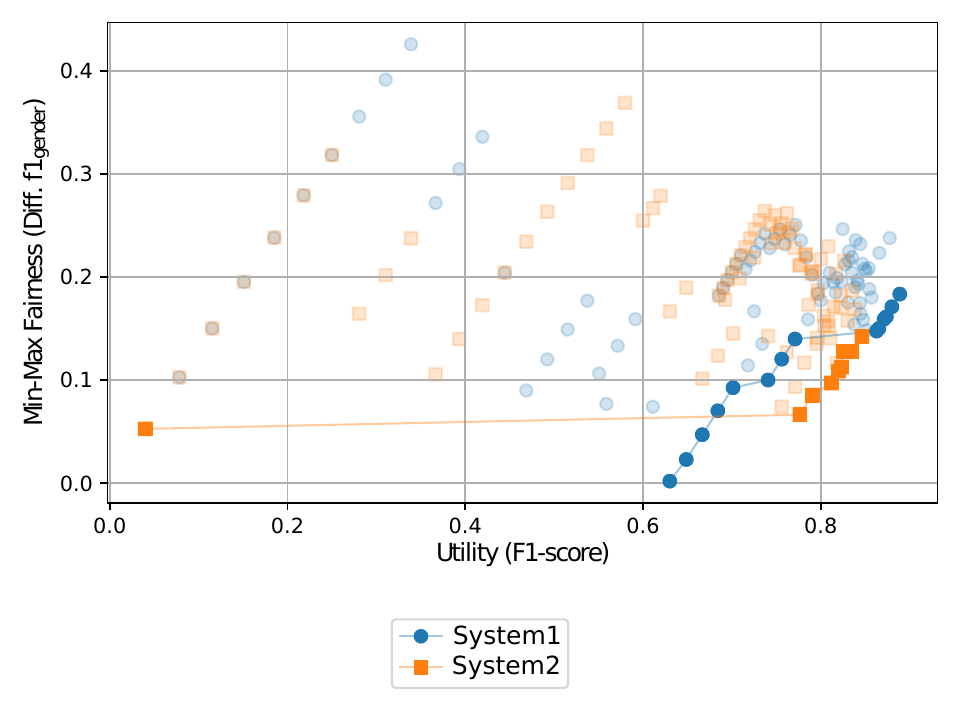}
        }
        \hspace{0.3cm}
        \subfloat[Radar chart.\label{fig:radar_shenzhen_gender}]{
            \includegraphics[width=0.80\columnwidth]{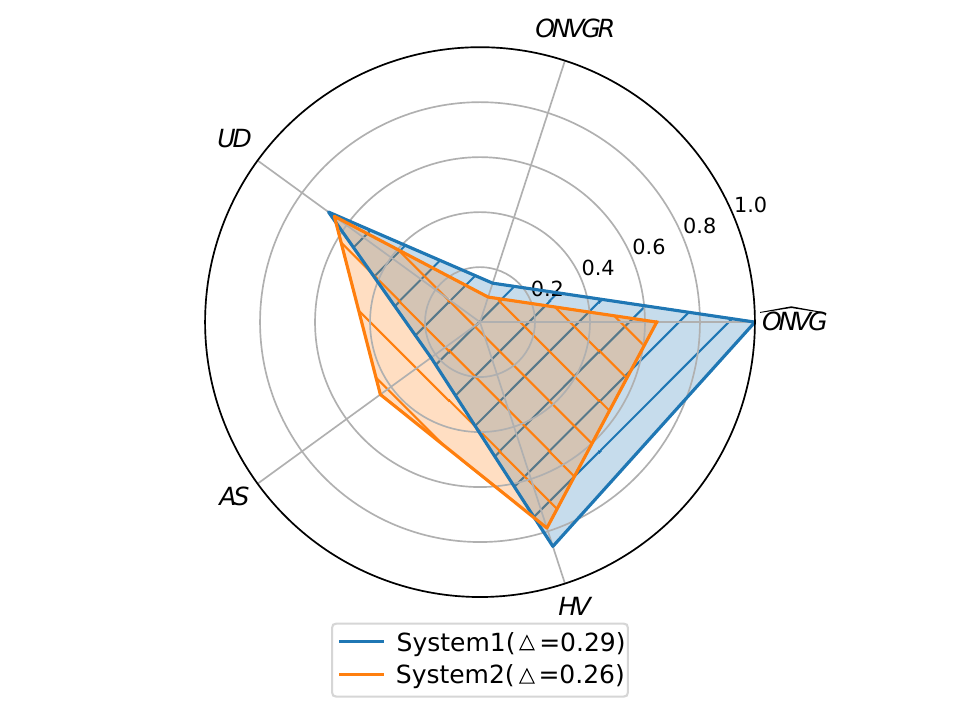}
        }
    \end{center}
    \subfloat[Quantitative results.\label{tab:radar_shenzhen_gender}]{
        \resizebox{0.95\columnwidth}{!}{
            \begin{tabular}{l|c|c|c|c|c|c}
                \textbf{ }       & \textbf{Convergence-Diversity} & \multicolumn{2}{c|}{\textbf{Capacity}} & \multicolumn{2}{c|}{\textbf{Diversity}} & \textbf{ }                                                        \\ \hline
                \textbf{ }       & \textbf{ }                     & \multicolumn{2}{c|}{\textbf{ }}        & \textbf{Distribution}                   & \textbf{Spread}    & \textbf{ }                                   \\ \hline
                \textbf{System}  & $\bm{\mathit{HV}}$             & $\bm{\widehat{\mathit{ONVG}}}$         & $\bm{\mathit{ONVGR}}$                   & $\bm{\mathit{UD}}$ & $\bm{\mathit{AS}}$ & $\bm{\widehat{\Delta}}$ \\ \hline
                \textit{System1} & 0.86                           & 1.00                                   & 0.15                                    & 0.68               & 0.22               & 0.29                    \\
                \textit{System2} & 0.79                           & 0.64                                   & 0.10                                    & 0.65               & 0.45               & 0.26                    \\ \hline
            \end{tabular}
        }
    }
    \caption{DenseNet/LoRA-ViT-Small on Shenzhen.}
    \label{fig:shenzhen_gender}
\end{figure}
%
The comparative evaluation of the DenseNet- and LoRA-ViT-Small-based systems on the
mBRSET dataset is illustrated in Fig.~\ref{fig:mbrset_obesity}, which reports the
distribution of utility (F1-score) and fairness outcomes (equalized odds difference
(EOD) for the obesity attribute) across all ML models in the trade-off system.
We note that group-wise comparisons based on prevalence-sensitive metrics such as
F1-score should be interpreted with caution, as they are influenced by the underlying
base rate (class ratio) and may partially reflect distributional characteristics rather
than purely model-induced disparities.
The curves show that \textit{System2} has lower EOD scores for a broader range of
utility levels, indicating improved fairness consistency under varying preference
settings. Table~\ref{tab:radar_mbrset_obesity} further summarizes the evaluation for
PF. \textit{System2} achieves higher scores in both $\widehat{\mathit{ONVG}}$ and
\textit{UD}. The \textit{HV} metric also favors \textit{System2} ($0.70$ vs.~$0.64$),
reflecting a closer alignment with the \emph{ideal} reference point. In contrast,
\textit{System1} exhibits a higher \textit{ONVGR} and \textit{AS} values. When
aggregated using the area score $\widehat{\Delta}$, \textit{System2} achieves $0.44$
compared to $0.40$ for \textit{System1}, confirming that \textit{System2} exhibits a
higher overall trade-off performance structure when balancing diagnostic utility and
\emph{obesity fairness} on this dataset.

\begin{figure*}[ht]
    \centering
    \subfloat[3D plot in metric space of ML system performance for one utility metric (accuracy) and two fairness criteria (gender and race).\label{fig:empirical_3d}]{
        \centering
        \includegraphics[width=0.45\textwidth]{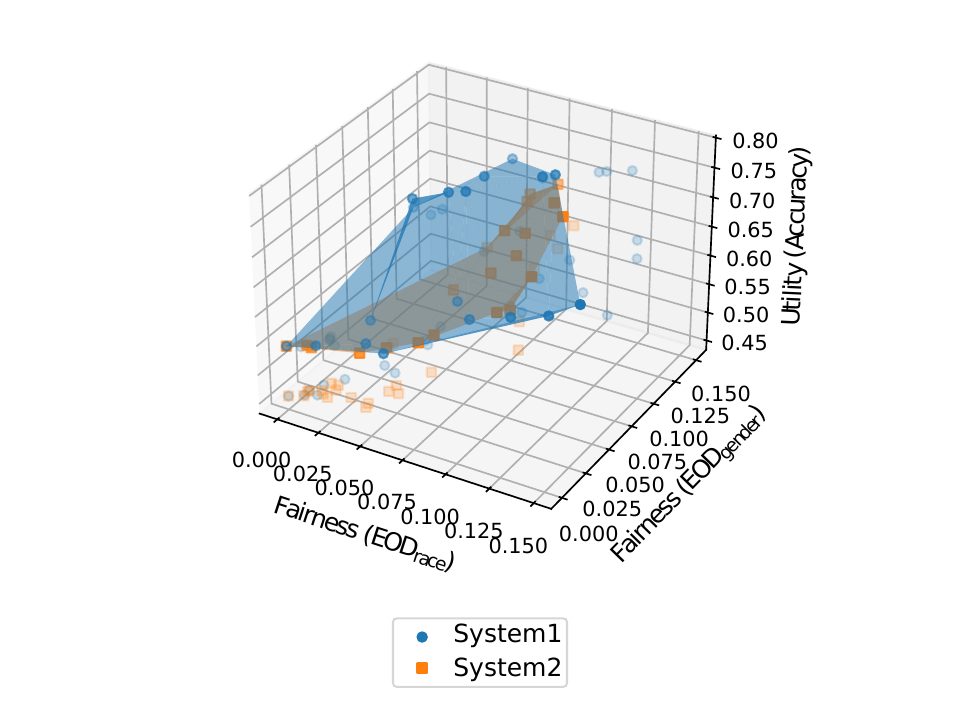}
    }
    \hspace{1cm}
    \subfloat[\textit{System2} (orange) slightly dominates \textit{System1} (blue) with a margin.\label{fig:empirical_radar}]{
        \centering
        \includegraphics[width=0.40\textwidth]{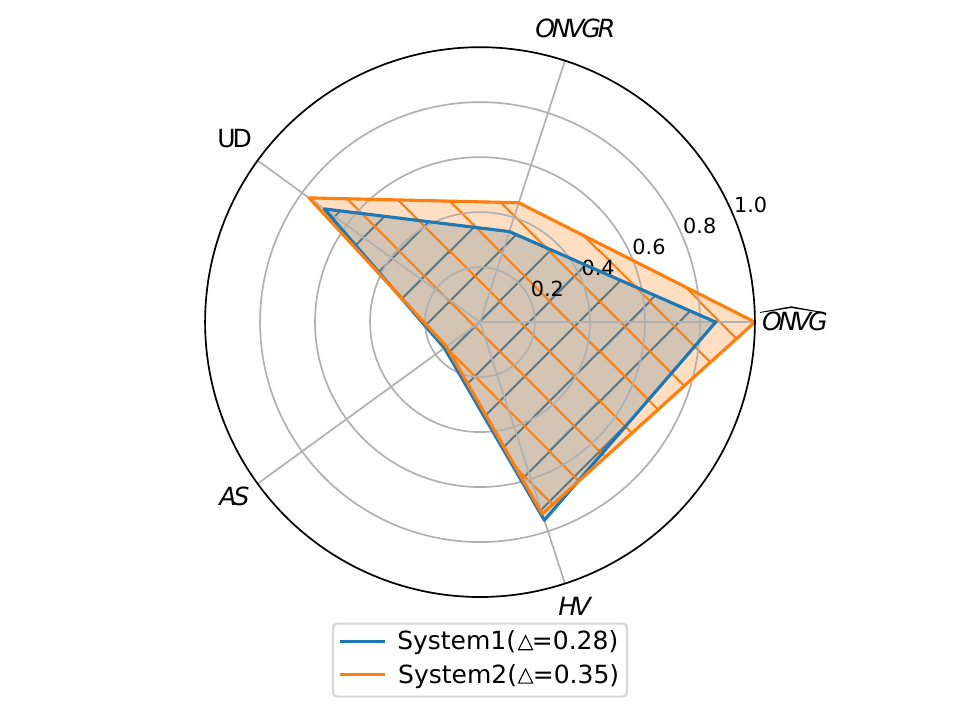}
    }
    \vspace{1cm}
    \subfloat[Quantitative results.\label{tab:empirical_whitebox}]{
        \resizebox{1.0\textwidth}{!}{
            \begin{tabular}{|l||c|c|c|c|c|c||c|c|c|c|c|c|c|c|c|c|c|c|}
                \hline
                \textbf{ }       & \textbf{Convergence-Diversity} & \multicolumn{2}{c|}{\textbf{Capacity}} & \multicolumn{2}{c|}{\textbf{Diversity}} & \textbf{ }         & \multicolumn{4}{c|}{\textbf{Model 1}} & \multicolumn{4}{c|}{\textbf{Model 10}} & \multicolumn{4}{c|}{\textbf{Model 25}}                                                                                                                                                                                                                                                                                                                           \\ \hline
                \textbf{ }       & \textbf{ }                     & \multicolumn{2}{c|}{\textbf{ }}        & \textbf{Distribution}                   & \textbf{Spread}    & \textbf{ }                            & \multicolumn{2}{c|}{\textbf{Utility}}  & \multicolumn{2}{c|}{\textbf{Fairness}} & \multicolumn{2}{c|}{\textbf{Utility}} & \multicolumn{2}{c|}{\textbf{Fairness}} & \multicolumn{2}{c|}{\textbf{Utility}} & \multicolumn{2}{c|}{\textbf{Fairness}}                                                                                                                                                         \\ \hline
                \textbf{System}  & $\bm{\mathit{HV}}$             & $\bm{\widehat{\mathit{ONVG}}}$         & $\bm{\mathit{ONVGR}}$                   & $\bm{\mathit{UD}}$ & $\bm{\mathit{AS}}$                    & $\bm{\widehat{\Delta}}$                & $\bm{\mathit{Acc}}$                    & $\bm{\mathit{F1}}$                    & $\bm{\mathit{DP}}$                     & $\bm{\mathit{EOdd}}$                  & $\bm{\mathit{Acc}}$                    & $\bm{\mathit{F1}}$ & $\bm{\mathit{DP}}$ & $\bm{\mathit{EOdd}}$ & $\bm{\mathit{Acc}}$ & $\bm{\mathit{F1}}$ & $\bm{\mathit{DP}}$ & $\bm{\mathit{EOdd}}$ \\ \hline
                \textit{System1} & 0.76                           & 0.86                                   & 0.35                                    & 0.70               & 0.16                                  & 0.28                                   & 0.78                                   & 0.77                                  & 0.07                                   & 0.09                                  & 0.68                                   & 0.75               & 0.06               & 0.08                 & 0.54                & 0.70               & 0.00               & 0.00                 \\
                \textit{System2} & 0.73                           & 1.00                                   & 0.46                                    & 0.77               & 0.15                                  & 0.35                                   & 0.76                                   & 0.76                                  & 0.08                                   & 0.10                                  & 0.56                                   & 0.71               & 0.02               & 0.05                 & 0.54                & 0.70               & 0.00               & 0.00                 \\ \hline
            \end{tabular}
        }
    }
    \caption{PHN on HGF.}
    \label{fig:empirical_whitebox}
\end{figure*}
%

The performance of the DenseNet- and LoRA-ViT-Small-based systems on Shenzhen dataset
is summarized in Fig.~\ref{fig:shenzhen_gender} that visualizes the utility performance
(F1-score) and fairness disparity (min-max difference for the gender groups). As shown
in the plots, \textit{System1} demonstrates a more favorable balance between fairness
variability and utility consistency. Specifically, \textit{System1} maintains a higher
performance on the F1-score without hitting very low scores while achieving comparable
EOD values against \textit{System2}. A quantitative comparison using PF-based
indicators is provided in Table~\ref{tab:radar_shenzhen_gender}. \textit{System1}
achieves higher performance in $\widehat{\mathit{ONVG}}$ with \textit{ONVGR}, and
outperforms \textit{System2} in \textit{UD} and \textit{HV} ($0.86$ vs.~$0.79$). While
\textit{System2} exhibits a larger \textit{AS} indicating a broader spread of trade-off
solutions, its distribution characteristic is weaker. Aggregating all indicators
through the area score shows \textit{System1} achieving $0.29$ compared to $0.26$ for
\textit{System2}, validating that \textit{System1} provides a higher trade-off
structure for \emph{gender fairness} in the Shenzhen dataset.

For the HGF experiment, Fig.~\ref{fig:empirical_whitebox} summarizes the evaluation of
the two PHN-based systems considering one utility performance (accuracy) and two
fairness metrics (EOD for the gender and race). Visual inspection alone illustrated by
the 3D performance plot in Fig.~\ref{fig:empirical_3d} is insufficient to reliably
differentiate the systems when multiple fairness criteria are considered alongside
utility. At this point, the radar visualization in Fig.~\ref{fig:empirical_radar}
provides a clearer qualitative aspect by showing a slight performance improvement for
\textit{System2} over \textit{System1}. This observation is further confirmed by the
quantitative indicators reported in Table~\ref{tab:empirical_whitebox}. Although there
is an inconsistency between \textit{HV} and the other performance indicators
($\widehat{\mathit{ONVG}}$, \textit{ONVGR} and \textit{UD}), an area of $0.35$ over
$0.28$ shows that \textit{System2} constitutes the better-performing utility-fairness
trade-offs than \textit{System1}. In this case, it is not possible to differentiate the
systems over \textit{AS} as they both exhibit average spreads of $0.15$ and $0.16$.
Finally, it is straightforward to interpret that both systems are far from optimality
as they are well below the ideal performance of $\widehat{\Delta}=1.00$.

Table~\ref{tab:empirical_whitebox} also reports the performance of individual sub-NNs
generated by the PHN (Models 1, 10, and 25), providing concrete examples of model-level
behavior within each utility-fairness trade-off system. These examples illustrate how
operating points guided by preference along the PF can differ substantially in their
balance between utility and fairness. For instance, in \textit{System1} (\emph{gender
    fairness}), the first sub-model achieves high utility (accuracy $\mathit{(Acc)}=0.78$
and F1-score $\mathit{(F1)}=0.77$), but exhibits demographic disparity (demographic
parity difference $\mathit{(DP)}=0.07$ and equalized odds difference
$\mathit{(EOdd)}=0.09$), whereas later models, such as 25th sub-NN, perform
substantially reduced unfairness ($\mathit{DP}=0.00$ and $\mathit{EOdd}=0.00$) at the
cost of lower accuracy and F1-score ($\mathit{Acc}=0.54$ and $\mathit{F1}=0.70$). A
similar trend is observed for \textit{System2} (\emph{race fairness}), where earlier
preference vectors favor utility while later ones build more equitable outcomes across
racial groups. However, evaluating these systems solely through individual model
outputs is insufficient, as meaningful assessment requires understanding their behavior
under multiple fairness criteria. Although \textit{System1} and \textit{System2} are
trained and evaluated on gender and race fairness, respectively, a complete analysis
also demands examining cross-criterion disparities to capture the broader fairness
landscape. These examples highlight how the proposed framework supports aggregated,
system-level interpretation rather than relying on isolated operating points.
Ultimately, this reinforces that fairness-aware evaluation must adopt a multi-objective
perspective that jointly reflects utility and multiple fairness criteria.

\section{Discussion}
\label{sec:discussion}
This work presents the requirements and advantages of an evaluation framework assessing
multidimensional utility-fairness trade-offs obtained with ML systems. The framework
enhances the comparison of different modeling strategies, with the goal of selecting
optimal solutions for real-world applications that require the assessment of multiple
fairness criteria. The ease of use and effectiveness of the proposed framework are
explored through comprehensive simulations and empirical studies using medical imaging
datasets designed to test fairness optimization approaches. Unlike previous evaluation
frameworks, this work provides a series of steps for the selection process of ML
systems in the context of multidimensional fairness exploring different criteria,
supported by MOO principles. Characterizing the optimal PF is particularly useful in
tasks where contradictory fairness performance indicators cannot be avoided, and
trade-offs should be tuned to specific fairness requirements from decision-makers. Our
framework could guide the tuning of ML systems to different notions of fairness for any
sensitive attribute and metric in a simple and transparent manner.
Fairness is interpreted as a multidimensional evaluation space, and the spectrum of
trade-offs is captured across demographic attributes rather than enforcing a single
optimal fairness configuration, which may fail to represent the overall capability of a
fairness-aware ML system.

While the utility-fairness trade-off is commonly reported in fairness-aware ML studies,
it does not universally manifest across all settings. As discussed in previous
works~\citep{chouldechova2017fair,kleinberg2016inherent}, different fairness
definitions such as \emph{independence}, \emph{separation}, and \emph{sufficiency} are
mutually incompatible under certain statistical assumptions, which means that
optimizing multiple fairness notions simultaneously is impossible to achieve.
Furthermore, recent works also show that the perceived utility-fairness tension may not
always exist when dataset-level biases are
addressed~\citep{wick2019unlocking,dutta2020there}. Conversely, a more recent work
by~\cite{dehdashtian2024utility} suggests that utility-fairness trade-offs can indeed
arise from intrinsic dataset characteristics by empirically observing the persistence
of such tension. Taken together, these findings indicate that utility-fairness
interactions are highly context dependent and influenced by both data distribution and
model assumptions. In this ongoing research landscape, our model- and metric-agnostic
evaluation framework contributes by providing a structured way to \emph{capture a
    snapshot} of how different methods achieve the balance/compensation between utility and
fairness. By quantifying these dynamics in a unified evaluation space, the framework
allows for a systematic comparison of fairness-aware ML systems independent of their
underlying architectures or optimization strategies, and it provides a baseline for
future work that needs a benchmark for multi-objective fairness evaluation. Moreover,
the framework is designed flexibly for practitioners so that they can select and
evaluate the specific utility-fairness trade-offs relevant to their application, rather
than imposing any pre-determined balance among objectives or assuming a universal
trade-off structure.

As mentioned in Section~\ref{sec:background}, the comprehensive analysis of PFs in
multiple dimensions is a limitation of previously proposed fairness evaluation
approaches. A summarized objective representation of performance indicators from the
PFs, both as a radar chart and as a measurement table, overcomes the limitations from
visualizing performance plots in multiple dimensions resulting from the assessment of
different ML systems. Furthermore, analogous to AUC over ROC, our evaluation framework
can transform the qualitative trend into a quantitative measurement, thus gathering all
necessary information for the interpretation of the performance gap between the
trade-off systems in consideration and against an ideal solution.

Beyond the medical imaging domain, the proposed framework has broader implications for
fairness evaluation across ML applications, as our approach generalizes the evaluation
process by integrating multiple fairness and utility criteria within a single
multi-objective formulation, independent of the domain under consideration. This
generalization enables the framework to serve as a common evaluation protocol for
comparing utility-fairness trade-offs in different contexts, such as financial risk
assessment and biometric systems. In this regard, our results complement existing
fairness benchmarking protocols by providing a structured, quantitative, and
qualitative means of summarizing utility-fairness trade-off performance across domains.

There may be some limitations of the proposed evaluation framework. A restriction is
the exponential cost for computing the MOO-based performance indicators as the number
of objectives increases~\citep{audet2021performance}. This scalability challenge is
particularly relevant when fairness must be evaluated across many demographic axes or
multiple fairness notions simultaneously, which may arise in complex real-world
deployments. However, we argue that the majority of tasks studied in the literature
focus on a small set of sensitive attributes, \ie, gender, age, and race. The use of
alternative performance indicators, such as Inverted Generational Distance
(\textit{IGD})~\citep{coello2004study}, could be explored to improve efficiency.
Another issue to consider is the equal weighting of all performance indicators, which
normalizes the contribution of the different indicators in the final evaluation. This
may not be desirable in situations where an indicator can evaluate the systems in
suboptimal performance and needs to have less impact on the final decision compared to
others. The proposed evaluation framework could be extended to support the dynamic
weighting of the indicators, so that the re-weighting can be performed based on the use
case. As another limitation, there may also be situations in which the indicator
measurements are the same or not feasible (refer to \textit{AS} and \textit{UD} in
Fig.~\ref{fig:syndata_point_point}) for all systems in comparison. In this case, such
an indicator is not informative for the final decision and the proposed framework
simplifies the evaluation to the joint contribution of the rest. This issue could be
alleviated by using different indicators that discriminate more, as the proposed
framework supports seamlessly including/excluding different types of indicators.
Finally, as emphasized by~\cite{selbst2019fairness}, algorithmic modeling and
evaluation of fairness cannot be fully abstracted from its social context. Accordingly,
the proposed evaluation framework should be interpreted as an assessment tool for
examining utility-fairness trade-offs under clearly defined assumptions and
constraints, rather than as a universal fairness solution applicable to all real-world
scenarios.

\section{Conclusions}
\label{sec:conclusion}
This paper proposes a multi-objective evaluation framework for utility-fairness
trade-offs resulting from ML systems, using performance indicators based on MOO. The
proposed framework is model- and task-agnostic, allowing for high flexibility in the
comparison of ML strategies, even when they have been optimized for different
objectives. This is an adaptive assessment framework that supports any kind of
performance indicators, including the proposed method, for convergence, diversity and
capacity analysis. The proposed framework is able to perform a comprehensive analysis
of ML systems with a measurement table and radar chart, overcoming the limitations
resulting from the qualitative assessment of solutions with multiple fairness
requirements. These tools provide a structured and visual means to evaluate and compare
multiple fairness metrics in ML systems. The measurement table allows for a clear,
organized presentation of the data, while the radar chart offers a visual
representation of how well the system performs across various utility and fairness
criteria using MOO indicators. The effectiveness of the evaluation approach is verified
by performing simulations and empirical analyses for a variety of use cases, with both
black- and white-box ML systems.
In particular, the empirical results on medical imaging based use cases illustrate how
the framework can expose fairness disparities between diagnostic models, guiding the
selection of appropriate trade-offs for ML systems in healthcare applications. The
proposed system is made available for public access to be applied in the context of
multi-objective evaluation for any domain.


\acks{This work was supported by the Swiss National Science Foundation (SNSF) through the project FairMI - Machine Learning Fairness with Application to Medical Images under grant number 214653. We also thank Fundação de Amparo à Pesquisa do Estado de São Paulo (FAPESP), grants 21/14725-3 and 2023/12468-9.}

%
\ethics{The work follows appropriate ethical standards in conducting research and writing the manuscript, following all applicable laws and regulations regarding treatment of animals or human subjects.}

\coi{We declare we don't have conflicts of interest.}

\data{All datasets used in this study are publicly available:

    Harvard Glaucoma Fairness (HGF):
    \url{https://github.com/Harvard-Ophthalmology-AI-Lab/Harvard-GF};

    Shenzhen Chest X-ray:
    \url{https://data.lhncbc.nlm.nih.gov/public/Tuberculosis-Chest-X-ray-Datasets/Shenzhen-Hospital-CXR-Set/index.html};

    mBRSET: \url{https://physionet.org/content/mbrset/1.0/}. }

\bibliography{references}

@inproceedings{zitzler1998multiobjective,
  title={Multiobjective optimization using evolutionary algorithms—a comparative case study},
  author={Zitzler, Eckart and Thiele, Lothar},
  booktitle={International conference on parallel problem solving from nature},
  pages={292--301},
  year={1998},
  organization={Springer}
}

@article{dosovitskiy2020image,
  title={An image is worth 16x16 words: Transformers for image recognition at scale},
  author={Dosovitskiy, Alexey},
  journal={arXiv preprint arXiv:2010.11929},
  year={2020}
}

@article{hu2022lora,
  title={Lora: Low-rank adaptation of large language models.},
  author={Hu, Edward J and Shen, Yelong and Wallis, Phillip and Allen-Zhu, Zeyuan and Li, Yuanzhi and Wang, Shean and Wang, Lu and Chen, Weizhu and others},
  journal={ICLR},
  volume={1},
  number={2},
  pages={3},
  year={2022}
}

@inproceedings{huang2017densely,
  title={Densely connected convolutional networks},
  author={Huang, Gao and Liu, Zhuang and Van Der Maaten, Laurens and Weinberger, Kilian Q},
  booktitle={Proceedings of the IEEE conference on computer vision and pattern recognition},
  pages={4700--4708},
  year={2017}
}

@article{jaeger2014two,
  title={Two public chest X-ray datasets for computer-aided screening of pulmonary diseases},
  author={Jaeger, Stefan and Candemir, Sema and Antani, Sameer and W{\'a}ng, Y{\`\i}-Xi{\'a}ng J and Lu, Pu-Xuan and Thoma, George},
  journal={Quantitative imaging in medicine and surgery},
  volume={4},
  number={6},
  pages={475},
  year={2014}
}

@article{schubert2017dbscan,
  title={DBSCAN revisited, revisited: why and how you should (still) use DBSCAN},
  author={Schubert, Erich and Sander, J{\"o}rg and Ester, Martin and Kriegel, Hans Peter and Xu, Xiaowei},
  journal={ACM Transactions on Database Systems (TODS)},
  volume={42},
  number={3},
  pages={1--21},
  year={2017},
  publisher={Acm New York, NY, USA}
}

@article{liu2022accuracy,
  title={Accuracy and fairness trade-offs in machine learning: A stochastic multi-objective approach},
  author={Liu, Suyun and Vicente, Luis Nunes},
  journal={Computational Management Science},
  volume={19},
  number={3},
  pages={513--537},
  year={2022},
  publisher={Springer}
}

@inproceedings{gustafson2023facet,
  title={Facet: Fairness in computer vision evaluation benchmark},
  author={Gustafson, Laura and Rolland, Chloe and Ravi, Nikhila and Duval, Quentin and Adcock, Aaron and Fu, Cheng-Yang and Hall, Melissa and Ross, Candace},
  booktitle={Proceedings of the IEEE/CVF international conference on computer vision},
  pages={20370--20382},
  year={2023}
}

@article{weerts2023fairlearn,
  title={Fairlearn: Assessing and improving fairness of ai systems},
  author={Weerts, Hilde and Dud{\'\i}k, Miroslav and Edgar, Richard and Jalali, Adrin and Lutz, Roman and Madaio, Michael},
  journal={Journal of Machine Learning Research},
  volume={24},
  number={257},
  pages={1--8},
  year={2023}
}

@article{garin2023medical,
  title={Medical imaging data science competitions should report dataset demographics and evaluate for bias},
  author={Garin, Sean P and Parekh, Vishwa S and Sulam, Jeremias and Yi, Paul H},
  journal={Nature medicine},
  volume={29},
  number={5},
  pages={1038--1039},
  year={2023},
  publisher={Nature Publishing Group US New York}
}

@inproceedings{agarwal2018reductions,
  title={A reductions approach to fair classification},
  author={Agarwal, Alekh and Beygelzimer, Alina and Dud{\'\i}k, Miroslav and Langford, John and Wallach, Hanna},
  booktitle={International conference on machine learning},
  pages={60--69},
  year={2018},
  organization={PMLR}
}

@article{kuratomi2025subgroup,
  title={Subgroup fairness based on shared counterfactuals},
  author={Kuratomi, Alejandro and Lee, Zed and Tsaparas, Panayiotis and Pitoura, Evaggelia and Lindgren, Tony and Dinis Junior, Guilherme and Papapetrou, Panagiotis},
  journal={Knowledge and Information Systems},
  pages={1--39},
  year={2025},
  publisher={Springer}
}

@inproceedings{chan2024group,
  title={Group fairness via group consensus},
  author={Chan, Eunice and Liu, Zhining and Qiu, Ruizhong and Zhang, Yuheng and Maciejewski, Ross and Tong, Hanghang},
  booktitle={Proceedings of the 2024 ACM Conference on Fairness, Accountability, and Transparency},
  pages={1788--1808},
  year={2024}
}

@inproceedings{diana2021minimax,
  title={Minimax group fairness: Algorithms and experiments},
  author={Diana, Emily and Gill, Wesley and Kearns, Michael and Kenthapadi, Krishnaram and Roth, Aaron},
  booktitle={Proceedings of the 2021 AAAI/ACM Conference on AI, Ethics, and Society},
  pages={66--76},
  year={2021}
}

@inproceedings{dwork2012fairness,
  title={Fairness through awareness},
  author={Dwork, Cynthia and Hardt, Moritz and Pitassi, Toniann and Reingold, Omer and Zemel, Richard},
  booktitle={Proceedings of the 3rd innovations in theoretical computer science conference},
  pages={214--226},
  year={2012}
}

@article{petersen2021post,
  title={Post-processing for individual fairness},
  author={Petersen, Felix and Mukherjee, Debarghya and Sun, Yuekai and Yurochkin, Mikhail},
  journal={Advances in Neural Information Processing Systems},
  volume={34},
  pages={25944--25955},
  year={2021}
}

@inproceedings{zhang2024mitigating,
  title={Mitigating label bias in machine learning: Fairness through confident learning},
  author={Zhang, Yixuan and Li, Boyu and Ling, Zenan and Zhou, Feng},
  booktitle={Proceedings of the AAAI Conference on Artificial Intelligence},
  volume={38},
  number={15},
  pages={16917--16925},
  year={2024}
}

@article{chouldechova2017fair,
  title={Fair prediction with disparate impact: A study of bias in recidivism prediction instruments},
  author={Chouldechova, Alexandra},
  journal={Big data},
  volume={5},
  number={2},
  pages={153--163},
  year={2017},
  publisher={Mary Ann Liebert, Inc. 140 Huguenot Street, 3rd Floor New Rochelle, NY 10801 USA}
}

@article{kleinberg2016inherent,
  title={Inherent trade-offs in the fair determination of risk scores},
  author={Kleinberg, Jon and Mullainathan, Sendhil and Raghavan, Manish},
  journal={arXiv preprint arXiv:1609.05807},
  year={2016}
}

@article{wick2019unlocking,
  title={Unlocking fairness: a trade-off revisited},
  author={Wick, Michael and Tristan, Jean-Baptiste and others},
  journal={Advances in neural information processing systems},
  volume={32},
  year={2019}
}

@inproceedings{dutta2020there,
  title={Is there a trade-off between fairness and accuracy? a perspective using mismatched hypothesis testing},
  author={Dutta, Sanghamitra and Wei, Dennis and Yueksel, Hazar and Chen, Pin-Yu and Liu, Sijia and Varshney, Kush},
  booktitle={International conference on machine learning},
  pages={2803--2813},
  year={2020},
  organization={PMLR}
}

@inproceedings{dehdashtian2024utility,
  title={Utility-fairness trade-offs and how to find them},
  author={Dehdashtian, Sepehr and Sadeghi, Bashir and Boddeti, Vishnu Naresh},
  booktitle={Proceedings of the IEEE/CVF Conference on Computer Vision and Pattern Recognition},
  pages={12037--12046},
  year={2024}
}

@inproceedings{selbst2019fairness,
  title={Fairness and abstraction in sociotechnical systems},
  author={Selbst, Andrew D and Boyd, Danah and Friedler, Sorelle A and Venkatasubramanian, Suresh and Vertesi, Janet},
  booktitle={Proceedings of the conference on fairness, accountability, and transparency},
  pages={59--68},
  year={2019}
}

@article{wu2025portable,
  title={A portable retina fundus photos dataset for clinical, demographic, and diabetic retinopathy prediction},
  author={Wu, Chenwei and Restrepo, David and Nakayama, Luis Filipe and Zago Ribeiro, Lucas and Shuai, Zitao and Barboza, Nathan Santos and Sousa, Maria Luiza Vieira and Fitterman, Raul Dias and Pereira, Alexandre Durao Alves and Regatieri, Caio Vinicius Saito and others},
  journal={Scientific Data},
  volume={12},
  number={1},
  pages={323},
  year={2025},
  publisher={Nature Publishing Group UK London}
}

@article{lu2020multiobjective,
  title={Multiobjective evolutionary design of deep convolutional neural networks for image classification},
  author={Lu, Zhichao and Whalen, Ian and Dhebar, Yashesh and Deb, Kalyanmoy and Goodman, Erik D and Banzhaf, Wolfgang and Boddeti, Vishnu Naresh},
  journal={IEEE Transactions on Evolutionary Computation},
  volume={25},
  number={2},
  pages={277--291},
  year={2020},
  publisher={IEEE}
}

@article{akter2022glaucoma,
  title={Glaucoma diagnosis using multi-feature analysis and a deep learning technique},
  author={Akter, Nahida and Fletcher, John and Perry, Stuart and Simunovic, Matthew P and Briggs, Nancy and Roy, Maitreyee},
  journal={Scientific Reports},
  volume={12},
  number={1},
  pages={8064},
  year={2022},
  publisher={Nature Publishing Group UK London}
}

@inproceedings{wang2020towards,
  title={Towards fairness in visual recognition: Effective strategies for bias mitigation},
  author={Wang, Zeyu and Qinami, Klint and Karakozis, Ioannis Christos and Genova, Kyle and Nair, Prem and Hata, Kenji and Russakovsky, Olga},
  booktitle={Proceedings of the IEEE/CVF conference on computer vision and pattern recognition},
  pages={8919--8928},
  year={2020}
}

@inproceedings{jo2023learning,
  title={Learning optimal fair decision trees: Trade-offs between interpretability, fairness, and accuracy},
  author={Jo, Nathanael and Aghaei, Sina and Benson, Jack and Gomez, Andres and Vayanos, Phebe},
  booktitle={Proceedings of the 2023 AAAI/ACM Conference on AI, Ethics, and Society},
  pages={181--192},
  year={2023}
}

@inproceedings{jang2022group,
  title={Group-aware threshold adaptation for fair classification},
  author={Jang, Taeuk and Shi, Pengyi and Wang, Xiaoqian},
  booktitle={Proceedings of the AAAI Conference on Artificial Intelligence},
  volume={36},
  number={6},
  pages={6988--6995},
  year={2022}
}

@inproceedings{kim2020fact,
  title={Fact: A diagnostic for group fairness trade-offs},
  author={Kim, Joon Sik and Chen, Jiahao and Talwalkar, Ameet},
  booktitle={International Conference on Machine Learning},
  pages={5264--5274},
  year={2020},
  organization={PMLR}
}

@inproceedings{roy2019mitigating,
  title={Mitigating information leakage in image representations: A maximum entropy approach},
  author={Roy, Proteek Chandan and Boddeti, Vishnu Naresh},
  booktitle={Proceedings of the IEEE/CVF Conference on Computer Vision and Pattern Recognition},
  pages={2586--2594},
  year={2019}
}

@inproceedings{jovanovic2023fare,
  title={Fare: Provably fair representation learning with practical certificates},
  author={Jovanovi{\'c}, Nikola and Balunovic, Mislav and Dimitrov, Dimitar Iliev and Vechev, Martin},
  booktitle={International Conference on Machine Learning},
  pages={15401--15420},
  year={2023},
  organization={PMLR}
}

@article{lahoti2020fairness,
  title={Fairness without demographics through adversarially reweighted learning},
  author={Lahoti, Preethi and Beutel, Alex and Chen, Jilin and Lee, Kang and Prost, Flavien and Thain, Nithum and Wang, Xuezhi and Chi, Ed},
  journal={Advances in neural information processing systems},
  volume={33},
  pages={728--740},
  year={2020}
}

@inproceedings{liu2021just,
  title={Just train twice: Improving group robustness without training group information},
  author={Liu, Evan Z and Haghgoo, Behzad and Chen, Annie S and Raghunathan, Aditi and Koh, Pang Wei and Sagawa, Shiori and Liang, Percy and Finn, Chelsea},
  booktitle={International Conference on Machine Learning},
  pages={6781--6792},
  year={2021},
  organization={PMLR}
}

@inproceedings{jang2023difficulty,
  title={Difficulty-based sampling for debiased contrastive representation learning},
  author={Jang, Taeuk and Wang, Xiaoqian},
  booktitle={Proceedings of the IEEE/CVF Conference on Computer Vision and Pattern Recognition},
  pages={24039--24048},
  year={2023}
}

@article{hardt2016equality,
  title={Equality of opportunity in supervised learning},
  author={Hardt, Moritz and Price, Eric and Srebro, Nati},
  journal={Advances in neural information processing systems},
  volume={29},
  year={2016}
}

@inproceedings{padh2021addressing,
  title={Addressing fairness in classification with a model-agnostic multi-objective algorithm},
  author={Padh, Kirtan and Antognini, Diego and Lejal-Glaude, Emma and Faltings, Boi and Musat, Claudiu},
  booktitle={Uncertainty in artificial intelligence},
  pages={600--609},
  year={2021},
  organization={PMLR}
}

@inproceedings{zhang2021fairer,
  title={Fairer machine learning through multi-objective evolutionary learning},
  author={Zhang, Qingquan and Liu, Jialin and Zhang, Zeqi and Wen, Junyi and Mao, Bifei and Yao, Xin},
  booktitle={International conference on artificial neural networks},
  pages={111--123},
  year={2021},
  organization={Springer}
}

@inproceedings{coello2004study,
  title={A study of the parallelization of a coevolutionary multi-objective evolutionary algorithm},
  author={Coello Coello, Carlos A and Reyes Sierra, Margarita},
  booktitle={Mexican international conference on artificial intelligence},
  pages={688--697},
  year={2004},
  organization={Springer}
}

@article{braden1986surveyor,
  title={The surveyor's area formula},
  author={Braden, Bart},
  journal={The College Mathematics Journal},
  volume={17},
  number={4},
  pages={326--337},
  year={1986},
  publisher={Taylor \& Francis}
}

@article{khachatryan2019primary,
  title={Primary open-angle African American glaucoma genetics (POAAGG) study: gender and risk of POAG in African Americans},
  author={Khachatryan, Naira and Pistilli, Maxwell and Maguire, Maureen G and Salowe, Rebecca J and Fertig, Raymond M and Moore, Tanisha and Gudiseva, Harini V and Chavali, Venkata RM and Collins, David W and Daniel, Ebenezer and others},
  journal={PloS one},
  volume={14},
  number={8},
  pages={e0218804},
  year={2019},
  publisher={Public Library of Science San Francisco, CA USA}
}

@article{dutt2023fairtune,
  title={Fairtune: Optimizing parameter efficient fine tuning for fairness in medical image analysis},
  author={Dutt, Raman and Bohdal, Ondrej and Tsaftaris, Sotirios A and Hospedales, Timothy},
  journal={arXiv preprint arXiv:2310.05055},
  year={2023}
}

@inproceedings{buolamwini2018gender,
  title={Gender shades: Intersectional accuracy disparities in commercial gender classification},
  author={Buolamwini, Joy and Gebru, Timnit},
  booktitle={Conference on fairness, accountability and transparency},
  pages={77--91},
  year={2018},
  organization={PMLR}
}

@article{zhao2022inherent,
  title={Inherent tradeoffs in learning fair representations},
  author={Zhao, Han and Gordon, Geoffrey J},
  journal={Journal of Machine Learning Research},
  volume={23},
  number={57},
  pages={1--26},
  year={2022}
}

@article{tan2002evolutionary,
  title={Evolutionary algorithms for multi-objective optimization: Performance assessments and comparisons},
  author={Tan, Kay Chen and Lee, Tong Heng and Khor, Eik Fun},
  journal={Artificial intelligence review},
  volume={17},
  number={4},
  pages={251--290},
  year={2002},
  publisher={Springer}
}

@inproceedings{van2000measuring,
  title={On measuring multiobjective evolutionary algorithm performance},
  author={Van Veldhuizen, David A and Lamont, Gary B},
  booktitle={Proceedings of the 2000 congress on evolutionary computation. CEC00 (Cat. No. 00TH8512)},
  volume={1},
  pages={204--211},
  year={2000},
  organization={IEEE}
}

@article{ha2016hypernetworks,
  title={Hypernetworks},
  author={Ha, David and Dai, Andrew and Le, Quoc V},
  journal={arXiv preprint arXiv:1609.09106},
  year={2016}
}

@inproceedings{zietlow2022leveling,
  title={Leveling down in computer vision: Pareto inefficiencies in fair deep classifiers},
  author={Zietlow, Dominik and Lohaus, Michael and Balakrishnan, Guha and Kleindessner, Matth{\"a}us and Locatello, Francesco and Sch{\"o}lkopf, Bernhard and Russell, Chris},
  booktitle={Proceedings of the IEEE/CVF Conference on Computer Vision and Pattern Recognition},
  pages={10410--10421},
  year={2022}
}

@inproceedings{wang2021understanding,
  title={Understanding and improving fairness-accuracy trade-offs in multi-task learning},
  author={Wang, Yuyan and Wang, Xuezhi and Beutel, Alex and Prost, Flavien and Chen, Jilin and Chi, Ed H},
  booktitle={Proceedings of the 27th ACM SIGKDD Conference on Knowledge Discovery \& Data Mining},
  pages={1748--1757},
  year={2021}
}

@article{yang2024limits,
  title={The limits of fair medical imaging AI in real-world generalization},
  author={Yang, Yuzhe and Zhang, Haoran and Gichoya, Judy W and Katabi, Dina and Ghassemi, Marzyeh},
  journal={Nature Medicine},
  volume={30},
  number={10},
  pages={2838--2848},
  year={2024},
  publisher={Nature Publishing Group US New York}
}

@inproceedings{yang2023minimax,
  title={Minimax auc fairness: Efficient algorithm with provable convergence},
  author={Yang, Zhenhuan and Ko, Yan Lok and Varshney, Kush R and Ying, Yiming},
  booktitle={Proceedings of the AAAI Conference on Artificial Intelligence},
  volume={37},
  number={10},
  pages={11909--11917},
  year={2023}
}

@mastersthesis{little2023fairness,
  title={To the fairness frontier and beyond: Identifying, quantifying, and optimizing the fairness-accuracy pareto frontier},
  author={Little, Camille},
  year={2023},
  school={Rice University}
}

@article{wei2022fairness,
  title={The fairness-accuracy Pareto front},
  author={Wei, Susan and Niethammer, Marc},
  journal={Statistical Analysis and Data Mining: The ASA Data Science Journal},
  volume={15},
  number={3},
  pages={287--302},
  year={2022},
  publisher={Wiley Online Library}
}

@article{xinying2023guide,
  title={A guide to formulating fairness in an optimization model},
  author={Xinying Chen, Violet and Hooker, John N},
  journal={Annals of Operations Research},
  volume={326},
  number={1},
  pages={581--619},
  year={2023},
  publisher={Springer}
}

@article{luo2024harvard,
  title={Harvard glaucoma fairness: a retinal nerve disease dataset for fairness learning and fair identity normalization},
  author={Luo, Yan and Tian, Yu and Shi, Min and Pasquale, Louis R and Shen, Lucy Q and Zebardast, Nazlee and Elze, Tobias and Wang, Mengyu},
  journal={IEEE Transactions on Medical Imaging},
  volume={43},
  number={7},
  pages={2623--2633},
  year={2024},
  publisher={IEEE}
}

@article{jiang2014consistencies,
  title={Consistencies and contradictions of performance metrics in multiobjective optimization},
  author={Jiang, Siwei and Ong, Yew-Soon and Zhang, Jie and Feng, Liang},
  journal={IEEE transactions on cybernetics},
  volume={44},
  number={12},
  pages={2391--2404},
  year={2014},
  publisher={IEEE}
}

@article{zitzler2003performance,
  title={Performance assessment of multiobjective optimizers: An analysis and review},
  author={Zitzler, Eckart and Thiele, Lothar and Laumanns, Marco and Fonseca, Carlos M and Da Fonseca, Viviane Grunert},
  journal={IEEE Transactions on evolutionary computation},
  volume={7},
  number={2},
  pages={117--132},
  year={2003},
  publisher={IEEE}
}

@article{audet2021performance,
  title={Performance indicators in multiobjective optimization},
  author={Audet, Charles and Bigeon, Jean and Cartier, Dominique and Le Digabel, S{\'e}bastien and Salomon, Ludovic},
  journal={European journal of operational research},
  volume={292},
  number={2},
  pages={397--422},
  year={2021},
  publisher={Elsevier}
}

@article{wu2001metrics,
  title={Metrics for quality assessment of a multiobjective design optimization solution set},
  author={Wu, Jin and Azarm, Shapour},
  journal={J. Mech. Des.},
  volume={123},
  number={1},
  pages={18--25},
  year={2001}
}

@article{gong2023influence,
  title={Influence maximization considering fairness: A multi-objective optimization approach with prior knowledge},
  author={Gong, Hao and Guo, Chunxiang},
  journal={Expert Systems with Applications},
  volume={214},
  pages={119138},
  year={2023},
  publisher={Elsevier}
}

@book{barocas2023fairness,
  title={Fairness and machine learning: Limitations and opportunities},
  author={Barocas, Solon and Hardt, Moritz and Narayanan, Arvind},
  year={2023},
  publisher={MIT press}
}

@article{navon2020learning,
  title={Learning the pareto front with hypernetworks},
  author={Navon, Aviv and Shamsian, Aviv and Chechik, Gal and Fetaya, Ethan},
  journal={arXiv preprint arXiv:2010.04104},
  year={2020}
}

@article{castelnovo2022clarification,
  title={A clarification of the nuances in the fairness metrics landscape},
  author={Castelnovo, Alessandro and Crupi, Riccardo and Greco, Greta and Regoli, Daniele and Penco, Ilaria Giuseppina and Cosentini, Andrea Claudio},
  journal={Scientific reports},
  volume={12},
  number={1},
  pages={4209},
  year={2022},
  publisher={Nature Publishing Group UK London}
}

@article{rabonato2025systematic,
  title={A systematic review of fairness in machine learning},
  author={Rabonato, Ricardo Trainotti and Berton, Lilian},
  journal={AI and Ethics},
  volume={5},
  number={3},
  pages={1943--1954},
  year={2025},
  publisher={Springer}
}

@article{starke2022fairness,
  title={Fairness perceptions of algorithmic decision-making: A systematic review of the empirical literature},
  author={Starke, Christopher and Baleis, Janine and Keller, Birte and Marcinkowski, Frank},
  journal={Big Data \& Society},
  volume={9},
  number={2},
  pages={20539517221115189},
  year={2022},
  publisher={SAGE Publications Sage UK: London, England}
}

@article{pessach2022review,
  title={A review on fairness in machine learning},
  author={Pessach, Dana and Shmueli, Erez},
  journal={ACM Computing Surveys (CSUR)},
  volume={55},
  number={3},
  pages={1--44},
  year={2022},
  publisher={ACM New York, NY}
}


\clearpage

\end{document}